\documentclass{article}

\usepackage[preprint]{neurips_2026}

\usepackage[utf8]{inputenc} % allow utf-8 input
\usepackage[T1]{fontenc}    % use 8-bit T1 fonts
\usepackage{hyperref}       % hyperlinks
\usepackage{url}            % simple URL typesetting
\usepackage{booktabs}       % professional-quality tables
\usepackage{amsfonts}       % blackboard math symbols
\usepackage{nicefrac}       % compact symbols for 1/2, etc.
\usepackage{microtype}      % microtypography
\usepackage{xcolor}         % colors

\usepackage[inline]{enumitem}
\usepackage{graphicx}
\usepackage{url}
\usepackage{subcaption}
\usepackage{soul}
\usepackage{amsmath}
\usepackage{amsfonts}
\usepackage{tcolorbox}
\usepackage{booktabs}
\usepackage{multicol}
\usepackage{multirow}
\usepackage{algorithm}
\usepackage{algpseudocode}
\usepackage{colortbl}
\usepackage{amssymb}
\usepackage{wrapfig}
\usepackage{longtable}

\title{Attention-Augmented Multi-Agent Systems with Spatio-Temporal Relevance}

\title{Enhancing Multi-Agent Communication through Attention Steering with Context Relevance}

\author{%
  Hongxiang Zhang \\
  Purdue University \\
  West Lafayette, IN, USA \\
  \texttt{hxxzhang@purdue.edu} \\
  \And
  Yuan Tian \\
  Purdue University \\
  West Lafayette, IN, USA \\
  \texttt{tian211@purdue.edu} \\
  \AND
  Tianyi Zhang \\
  Purdue University \\
  West Lafayette, IN, USA \\
  \texttt{tianyi@purdue.edu} \\
}

\definecolor{lightblue}{RGB}{255, 255, 255}
\definecolor{lightgreen}{RGB}{210, 240, 210}
\definecolor{lightgray}{RGB}{225, 225, 225}
\definecolor{lightred}{rgb}{1.0, 0.8, 0.8}

\newcommand{\lightgreen}[1]{\sethlcolor{lightgreen}\hl{\textbf{#1}}}

\newcommand{\lightred}[1]{\sethlcolor{lightred}\hl{#1}}

\newcommand{\selfanchor}{\textsc{Agent-Radar}}
\newcommand{\headercolorlong}{\rowcolor{gray!15}}

\newcommand{\gain}[1]{\textcolor{green!50!black}{\scriptsize$_{\uparrow #1}$}}
\newcommand{\drop}[1]{\textcolor{red!70!black}{\scriptsize$_{\downarrow #1}$}}
\begin{document}

\maketitle

\begin{abstract}
LLM-based multi-agent systems have demonstrated remarkable performance on complex tasks through collaborative reasoning. However, these systems tend to rapidly accumulate extremely long conversation histories during interaction. As conversations lengthen, relevant information is increasingly diluted by irrelevant context, leading to degraded performance.  
In this work, we present {\selfanchor}, a training-free context management method that dynamically steers each agent's attention toward relevant context with a novel temporal and spatial decay mechanism. 
% toward relevant context extracted from the conversation history.
Our experiments demonstrate that {\selfanchor} outperforms state-of-the-art methods across five different benchmarks, yielding gains of up to 7.64 absolute points. Furthermore, our analysis shows that {\selfanchor} remains effective and robust as the number of agents and interaction rounds increases.  
Finally, the ablation study shows that core components in {\selfanchor} are crucial to performance and generalizable in different settings.
\end{abstract}

\section{Introduction}

% \begin{wrapfigure}{r}{0.48\columnwidth}
%     \vspace{-0.5em}
%     \centering
%     \includegraphics[width=0.46\columnwidth]{figure/context_grow.pdf}
%     \caption{The workflow of multi-agent systems. The agent communication often accumulates extremely long context windows rapidly.}
%     \label{fig:context_grow}
%     \vspace{-1em}
% \end{wrapfigure}

% Large Language Models (LLMs) have evolved beyond static text generators into autonomous agents capable of reasoning, interactive communication, and collaborative problem-solving~\citep{yao2022react, wang2024survey, liang2024encouraging}. Building on these capabilities, multi-agent systems (MAS) instantiate multiple LLM agents that demonstrate strong performance across a wide range of tasks, from multi-hop question answering to mathematical reasoning and complex problem solving~\citep{wang2024unleashing, du2024improving, jiang2023llm, wu2024autogen}. 
LLM-based multi-agent systems have demonstrated remarkable performance on complex tasks~\citep{liang2024encouraging, wang2024unleashing, du2024improving, wu2024autogen}. 
Despite these successes, agent communication often accumulates extremely long context windows rapidly~\citep{zhang2024cut} and suffers from communication redundancy~\citep{wang-etal-2025-agentdropout, zhang2024cut,du2024improving, liang2024encouraging}. As a result, the original task description, key constraints, and other critical evidence are often buried within the context, making it difficult for LLMs to reliably identify and attend to them. %receive insufficient attention. 
This phenomenon, also known as the ``lost-in-the-middle'' effect, frequently induces hallucinations and logical errors during generation~\citep{becker2026stay, liu-etal-2024-lost, cemri2025multi, zhang2025agent}. This issue is further exacerbated as systems scale. Increasing numbers of agents and interaction rounds cause communication histories to grow rapidly, amplifying context dilution and undermining the reliability of large-scale agent collaborations \citep{cemri2025multi,becker2026stay, zhang2025agent}.
% Prior work has sought to improve efficiency by pruning redundant agents or communication edges~\citep{wang-etal-2025-agentdropout}. While effective, these structural pruning techniques leave a critical gap: even after topology pruning, agents still reason over long, noisy, multi-round histories where attention dilution remains unresolved.

Prior context management methods have sought to mitigate this challenge in two ways. One line of work applies periodic summarization to compress accumulated interaction history~\citep{wang2025recursively,wu2025resum}. Another adopts pruning strategies that limit context growth by removing redundant agents or communication edges~\citep{zhang2024cut,wang-etal-2025-agentdropout}.
While effective at improving efficiency, these methods introduce information loss. Specifically, compressing or pruning messages can discard subtle but critical signals, potentially misleading downstream reasoning and propagating errors across subsequent interactions~\citep{shen2025understanding,acharjee2025remembers}. 

% Attention mechanism provides a natural alternative for context management. 
We propose a shift from context compression or pruning to selectively steering agents' attention over relevant context.
In the original Transformer architecture, the attention mechanism retrieves information by matching queries to keys, allowing the model to assign importance weights across preceding tokens in the context~\citep{vaswani2017attention}. Accordingly, in multi-agent systems, an agent's current generation step naturally acts as a query seeking to retrieve relevant information from the communication history. Unlike prior methods that act as hard, binary filters over context, we instead steer attention toward a subset of relevant content while preserving the full transcript, thereby avoiding information loss. %This perspective motivates a complementary question: \emph{how can we dynamically guide agents' attention toward relevant context without modifying the underlying context itself?}

\label{sec:intro}

To achieve this, we develop {\selfanchor}, a training-free method that dynamically guides agents' attention toward relevant context without modifying the underlying communication history. 
%Different from prior context management methods, {\selfanchor} preserves the original messages and communication topology. 
%{\selfanchor} is grounded in empirical evidence demonstrating that different messages in MAS contribute unequally~\citep{li-etal-2024-improving-multi, qian2024scaling}. 
% To operationalize this insight, {\selfanchor} modulates context across three distinct dimensions. 
To identify which parts of the history should guide each generation step, {\selfanchor} selects context by jointly modeling semantic relevance, spatial distance, and temporal recency.
First, inspired by organizational network theory~\citep{borgatti2009network, allen1984managing}, where influence decays with structural distance, {\selfanchor} formulates the agent interaction topology as a graph and applies a spatial decay factor based on inter-agent distance, prioritizing messages from closely interacting agents. 
% down-weighting messages that are structurally remote from the current agent. 
Second, {\selfanchor} incorporates temporal decay to down-weight stale information, reflecting evidence from both long-context modeling and human memory that relevance diminishes over time~\citep{finkenbinder1913curve, sun2023retentive, ma2022mega}. 
% \todo{earlier messages should not be treated as equally salient as recent evidence}
Third, {\selfanchor} performs fine-grained semantic matching by segmenting messages into sentence-level units to extract relevant evidence at the current step.
Finally, {\selfanchor} integrates these signals through a lightweight attention-steering mechanism, enabling agents to emphasize key instructions, critical constraints, and useful intermediate evidence, without compressing or pruning history tokens.

% Existing evidence indicates that different messages in MAS contribute unequally~\citep{li-etal-2024-improving-multi, qian2024scaling}; in organizational and social networks, information relevance and collaborative influence decay as structural distance increases~\citep{borgatti2009network, allen1984managing}; and both long-horizon sequence models and human-memory studies motivate temporal decay over stale context~\citep{finkenbinder1913curve,sun2023retentive,ma2022mega}. Starting from these insights, we build {\selfanchor} that dynamically steers agent attention along spatial proximity and temporal recency, while using semantic matching to identify task-critical content.

We evaluate {\selfanchor} on five benchmarks using three base LLMs. Compared with SOTA context management methods, including compression-based approach~\citep{wang2025recursively} and pruning methods~\citep{zhang2024cut,wang-etal-2025-agentdropout}, {\selfanchor} achieves gains of up to 7.64 absolute points.
As a plug-in method, {\selfanchor} can be seamlessly integrated into diverse multi-agent systems, including debate-based systems~\citep{du2024improving}, GPTSwarm~\citep{zhuge2024gptswarm}, and AutoGen~\citep{wu2024autogen}. Across these frameworks, {\selfanchor} consistently improves their performance by up to 12.87 absolute points and generalizes across different interaction topologies.
Furthermore, {\selfanchor} remains robust as systems scale to larger numbers of agents and longer interaction rounds. Finally, ablation studies confirm that each component contributes meaningfully to the overall performance gains.
\section{Problem Formulation}
    %\paragraph{Multi-Agent System as Graph.}
    \label{sec:formalization}
    We model the multi-agent system architecture as a graph  $\mathcal{G} = (\mathcal{V}, \mathcal{E}, \mathcal{F})$.
    Let $\mathcal{V} = \{v_i\}_{i=1}^N$ be the set of $N$ nodes, where each node $v_i$ represents an autonomous LLM agent. The edge set $\mathcal{E} \subseteq \mathcal{V} \times \mathcal{V}$ defines the communication topology. An edge $(v_i, v_j) \in \mathcal{E}$ denotes the communication path from the agent $v_i$ to the agent $v_j$. 
    The set of mapping functions $\mathcal{F} = \{f_i\}_{i=1}^N$ represents the inference processes of the agents. We further define the hop distance $d_{\mathcal{G}}(v_j, v_i)$ as the length of the shortest path from $v_j$ to $v_i$ in $\mathcal{G}$.
    The $k$-hop neighborhood of $v_i$ is $\mathcal{N}_{k}(v_i) = \{v_j \in \mathcal{V} \mid d_{\mathcal{G}}(v_j, v_i) \leq k\}$. All agents that  can reach to $v_i$ in $\mathcal{G}$ is $\mathcal{N}_{N-1}(v_i) = \bigcup_{k=1}^{N-1}\mathcal{N}_{k}(v_i)$. 

    The system proceeds over communication rounds $t \in \{1,\dots,T\}$. At time $t$, the active agent $v_i$ generates the response
    \begin{equation}
        o_i^{(t)} = f_i\!\left(q^{(t)},~\mathcal{Z}_i^{(t)}\right),
    \end{equation}
    where $q^{(t)}$ is the query received by the current agent at time $t$, and $\mathcal{Z}_i^{(t)}$ is the accumulated communication histories available to $v_i$. 
    The context available to the current agent $v_i$ at time $t$ is    
    \begin{equation}
        \mathcal{Z}_i^{(t)}=\{\ o_j^{(\tau)} \mid v_j \in \mathcal{N}_{N-1}(v_i),  ~1\le \tau < t\,\},
    \end{equation}
    where $o_j^{(\tau)}$ is the output produced by agent $v_j$ at timestamp $\tau$.

    In our method, we introduce an additional parameter $\hat{\mathcal{Z}}_i^{(t)}$ to $f_i$, denoting the selected context that {\selfanchor} steers attention towards for each inference step. In the following section, we describe how {\selfanchor} selects $\hat{\mathcal{Z}}_i^{(t)}$ from the accumulated communication histories and uses it as an additional argument to $f_i$ for attention steering.
    
    % \subsection{Context Growth}
    % The input prompt fed to $f_i^{(t)}$ is a serialization of $(q^{(t)}, \mathcal{Z}_i^{(t)})$. 
    % Consider a dense communication topology (e.g., fully connected) where every agent communicates with every other at every round and all historical responses are retained. 
    % In this scenario, the number of system-wide messages per round scales as $\Theta(N^2)$. Accumulated over $T$ rounds, the total message volume scales as $\Theta(N^2 T)$. 
    % For an individual receiver $v_i$, the retained history grows as $\Theta(NT)$ because it can collect responses from $\Theta(N)$ agents in each round. This growth directly motivates {\selfanchor} to prevent critical information from being overwhelmed by irrelevant history.
    
    % \todo{What dooes the subscript intra mean?}\todo{This is not fully correct. All agents connected to each other don't mean that they will necessarily talk to each other in each round. For example, in a decentralized agent architecture, every agent will decide who they will talk to next in the next round.}

\section{Methodology}
    \label{sec:methodology}
\begin{figure}
    \centering
    \includegraphics[width=\linewidth]{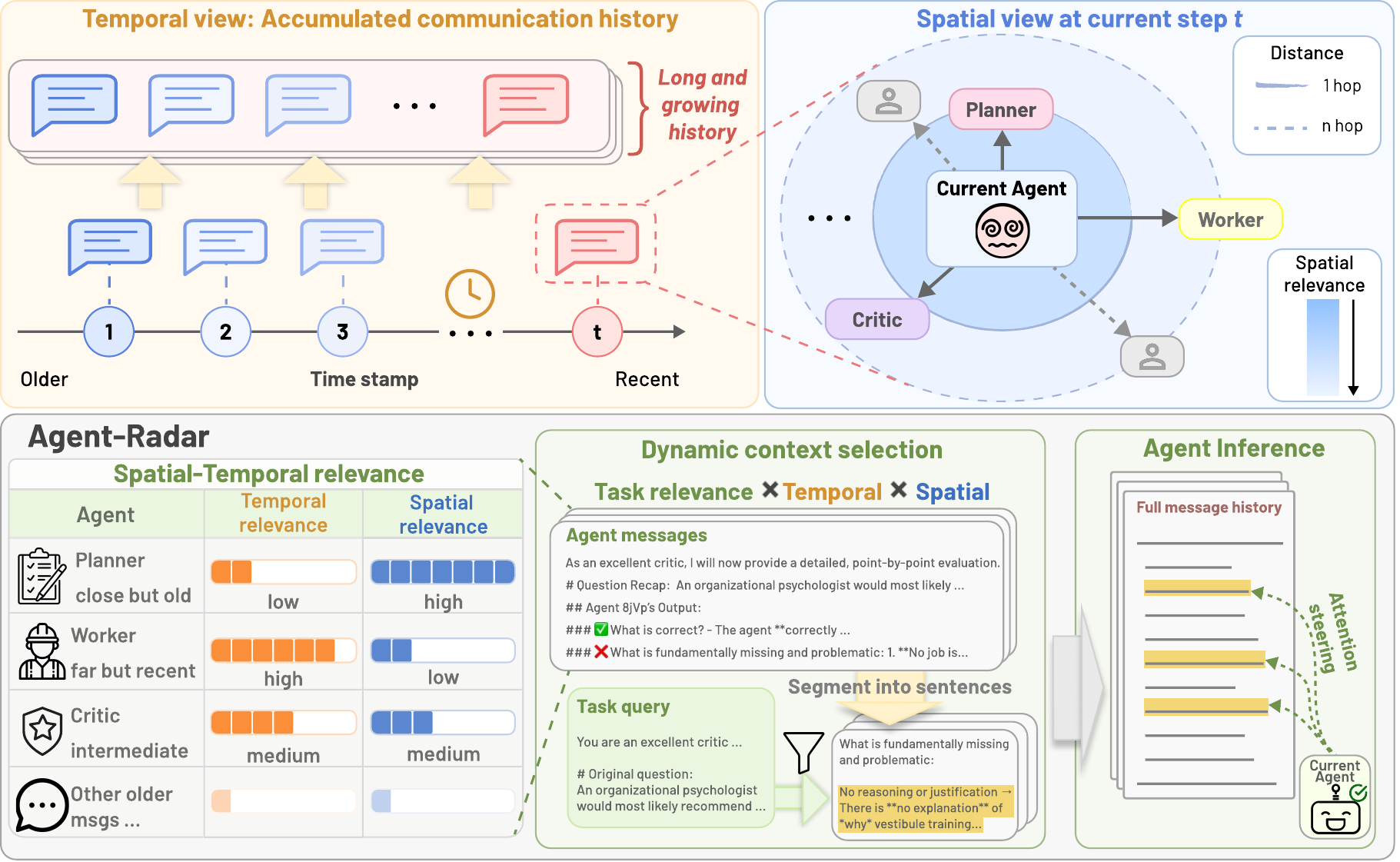}
    \caption{Overview of {\selfanchor}. \textbf{\textit{(Top)}} MAS interactions rapidly accumulate long communication histories, where useful information is buried in the middle, receiving insufficient attention. \textbf{\textit{(Bottom)}} {\selfanchor} preserves the full transcript and topology, scores sentence-level context by semantic relevance weighted with temporal and spatial decay, and steers the agent's attention toward the selected context during inference.}
    \vspace{-0.2cm}
    \label{fig:overview}
\end{figure}

    We introduce {\selfanchor}~(Figure~\ref{fig:overview}), a training-free method that augments multi-agent systems (MAS) by steering an agent's attention toward semantically relevant context that is both spatially proximate and temporally salient. Building on the formulation in Section~\ref{sec:formalization}, {\selfanchor} selects fine-grained context by jointly modeling semantic relevance, spatial structure, and temporal recency. Then, it steers the agent's attention toward the selected context during generation.

\subsection{Dynamic Context Selection}
    \label{sec:Dynamic Context Selection and Attention Steering}
    % Given a MAS graph $\mathcal{G}$ and the current agent under operation, $v_i$, at timestamp $t$, the raw message pool available to $v_i$ is
    % \begin{equation}
    %     \mathcal{C}_i^{(t)} =
    %         \bigl\{(v_j,\,\tau,\,o_j^{(\tau)}) \;\big|\;
    %                v_j \in \mathcal{N}_{N-1}(v_i),~o_j^{(\tau)} \text{ is selected for } v_i,\; \tau < t \bigr\}
    %         \;\cup\; \{q^{(t)}\},
    %     \label{eq:context_pool}
    % \end{equation}
    % where each triple $m_{j,\tau} \triangleq (v_j, \tau, o_j^{(\tau)})$ denotes a message originating from agent $v_j$ at time $\tau$. 

    % This pool includes direct neighbors and multi-hop connected agents. Without any management, the size of this message pool,  $|\mathcal{C}_i^{(t)}|$, grows at the rate of $\Theta\!\left(\left|\mathcal{N}_{N-1}(v_i)\right| \cdot t\right)$, progressively burying critical information among historical messages. The goal of {\selfanchor} is to identify the context relevant to the current inference objective of $v_i$ within this expanding communication history and steer its attention towards the relevant context during inference.
    
    While standard semantic retrieval is effective in many cases, applying it naively to the dynamic MAS context is insufficient.\footnote{We present ablation studies in Section~\ref{sec:ablation} comparing {\selfanchor} against semantic-only retrieval and variants with spatial or temporal decay individually removed.} 
    Unlike a flat document collection, multi-agent communication is generated by different agents at different rounds, and these source and temporal signals affect whether a previous response is likely to be useful for the current active agent. 
    In particular, a sentence from a previously generated output can remain semantically relevant to the inference objective of the current agent while being outdated, redundant, or weakly grounded. 
    Pure semantic retrieval ignores these signals and may therefore over-select stale or indirectly propagated evidence. 
    To this end, we formulate {\selfanchor} as retrieval over the spatio-temporal communication history. For each generated output $o_j^{(\tau)}$ in the communication history, we assign a scalar relevance score to each generated output that combines a spatial decay factor and a temporal decay factor:
    \begin{equation}
        r_i(o_j^{(\tau)} ;\, t)
        \;=\;
        \underbrace{\phi_s\!\bigl(v_j,\, v_i\bigr)}_{\text{spatial decay }}
        \;\cdot\;
        \underbrace{\phi_t\!\bigl(\tau;\, t\bigr)}_{\text{temporal decay } },
        \label{eq:scalar_score}
    \end{equation}
    where $\phi_s$ and $\phi_t$ are the spatial and temporal decay factors. Given these scores, we retrieve fine-grained, sentence-level attention-aligned context by combining this spatio-temporal score with semantic similarity to the query received by the current agent.

\paragraph{Spatial Decay.}
     The organizational network theory~\citep{borgatti2009network, allen1984managing} finds that information relevance in collaborative networks typically decays with structural distance. 
     In a multi-agent workflow, responses from nearby collaborators are more likely to contain direct, actionable evidence, whereas information propagating from distant agents often introduces indirect noise or redundant context. 
     Based on this insight, we use an exponential form because each additional relay can be viewed as retaining only a fixed fraction of the useful signal. If one extra hop retains a fraction $\lambda_s$ of the relevance, then $k$ extra hops retain $\lambda_s^k$.      
     Therefore, the first term in Equation~\ref{eq:scalar_score} models spatial relevance as a decay over graph distance:
    \begin{equation}
    \phi_s(v_j, v_i)
        = \lambda_s^{\, d_{\mathcal{G}}(v_j,\, v_i)-1},
        \quad 0 < \lambda_s < 1,
        \label{eq:spatial_decay}
    \end{equation}
    where $d_{\mathcal{G}}(v_j, v_i)$ is the distance from Section~\ref{sec:formalization} and $\lambda_s$ controls the decay rate. For direct neighbors or self-generated critiques, we set $\phi_s = 1$. Generated outputs from agents further away receive exponentially lower weights.
    
\paragraph{Temporal Decay.}
    As the discussion evolves, earlier responses should not be treated as equally informative as recent exchanges, because later rounds often refine, correct, or supersede earlier reasoning. This intuition is consistent with the recency effect observed in human memory~\citep{finkenbinder1913curve} and with temporal decay mechanisms in long-horizon sequence modeling~\citep{sun2023retentive, ma2022mega}. 
    We use the same idea for temporal decay that discounts responses by their age:
    \begin{equation}
        \phi_t(\tau;\, t)
        = \lambda_t ^{\!\bigl( t - \tau - 1\bigr)},
        \quad 0 < \lambda_t < 1,
        \label{eq:temporal_decay}
    \end{equation}
    where $\tau$ is the time when the response was produced, $t - \tau - 1 \ge 0$ is the age of the generated output, and $\lambda_t$ governs the decay rate. 
    The most recent response receives a weight of $1$, and older responses are discounted exponentially. 
    This keeps recent evidence significant while still allowing older but useful responses to contribute when they are semantically relevant.
    
\paragraph{Retrieval.}
    \label{sec:retrieval}
    On top of these two decay mechanisms, we perform fine-grained sentence-level retrieval. 
    Unlike previous methods that retrieve whole message chunks, steering attention to an entire generated response can still introduce irrelevant details, since only part of the message may be useful for the current query. We therefore retain only the relevant context from each generated output. 
    
    Specifically, we first segment each generated output  $o_j^{(\tau)}$ into a sequence of sentences $\mathcal{S}_{j,\tau} = \{s_k\}_{k=1}^{K_{j,\tau}}$. Each sentence $s_k$ is then encoded by a pre-trained sentence encoder $\mathrm{Enc}(\cdot)$\footnote{Sentence segmentation and retrieval are implemented using SentenceTransformer (\texttt{all-MiniLM-L6-v2} as the default encoder)~\citep{reimers-2019-sentence-bert}. Alternative retrieval methods are provided in Section~\ref{sec:analysis}.}:
    
    \begin{equation}
        \mathbf{e}_k = \mathrm{Enc}(s_k) \in \mathbb{R}^d,
        \quad
        \mathbf{e}_{q^{(t)}} = \mathrm{Enc}(q^{(t)}) \in \mathbb{R}^d.
        \label{eq:embedding}
    \end{equation}
    The semantic relevance of sentence $s_k$ to the current query $q^{(t)}$ is measured by cosine similarity:
    \begin{equation}
        \phi_{\mathrm{sem}}(s_k,\, q^{(t)})
        = \frac{\mathbf{e}_k^{\top} \mathbf{e}_{q^{(t)}}}
               {\|\mathbf{e}_k\|\,\|\mathbf{e}_{q^{(t)}}\|}
        \;\in [-1, 1].
        \label{eq:semantic}
    \end{equation}
    The final relevance score for sentence $s_k$ from generated output $o_j^{(\tau)}$ is obtained by combining the spatio-temporal scalar with semantic similarity:
    \begin{equation}
        \mathrm{score}(s_k \mid o_j^{(\tau)},\, v_i,\, t,\, q^{(t)})
        \;=\;
        r_i(o_j^{(\tau)};\, t)
        \;\cdot\;
        \phi_{\mathrm{sem}}(s_k,\, q^{(t)}).
        \label{eq:final_score}
    \end{equation}
    This factored design ensures that a relevant sentence must be both contextually proximate (spatially and temporally) and semantically aligned with the query $q^{(t)}$ received by the current agent.
    
    The attention-aligned context is constructed by retaining all sentences whose combined relevance score exceeds a threshold $\theta$, with the current query $q^{(t)}$ included:
    \begin{equation}
        \hat{\mathcal{Z}}_i^{(t)}
        = \{q^{(t)}\}
        \;\cup\;
        \Bigl\{
            s_k \;\Big|\;
            \mathrm{score}(s_k \mid o_j^{(\tau)}, v_i, t, q^{(t)}) \ge \theta,\;
            o_j^{(\tau)} \in \mathcal{Z}_i^{(t)}
        \Bigr\},
        \label{eq:aligned_context}
    \end{equation}
    where $\theta$ is a hyperparameter\footnote{We conduct parameter selection analysis in Appendix~\ref{sec:threshold_analysis}} that controls the trade-off between context coverage and compression. 
    This retrieval does not prune the transcript. Instead, it only determines which context receives explicit attention steering. Note that the current query $q^{(t)}$ is always included in the attention-aligned context, $\hat{\mathcal{Z}}_i^{(t)}$, so the selected context from previous conversations does not outweigh the current inference objective during attention steering. 
    %In addition, the query $q^{(t)}$ is always retained, ensuring that the original instruction and task constraints are never displaced by subsequent communication history.

\subsection{Attention Steering and Agent Inference}
    The agent inference is then updated to
    \begin{equation}
        o_i^{(t)}
        = f_i\!\left(q^{(t)},\;\mathcal{Z}_i^{(t)},\;\hat{\mathcal{Z}}_i^{(t)}\right).
        \label{eq:aligned_inference}
    \end{equation}
    where $\hat{\mathcal{Z}}_i^{(t)}$ denotes the selected context that {\selfanchor} steers attention towards. 
    To implement the attention intervention, we adopt Selective Prompt Anchoring (SPA)~\citep{tian2024selective} as a lightweight backend attention steering mechanism that amplifies attention to the sentences in $\hat{\mathcal{Z}}_i^{(t)}$. This design is plug-in and training-free: it does not modify the backbone LLM weights nor the MAS topology, and can be integrated with different MAS systems. We discuss the alternative attention steering method in Section~\ref{sec:steeringablation}. We summarize the full algorithm in Appendix~\ref{appendix:algorithm}.
\section{Experiments}

\subsection{Experimental Setup}

\paragraph{Benchmarks.} 
\label{sec:benchmarks}
We evaluate {\selfanchor} on three categories of benchmarks. For open-ended question answering, we use \textbf{HotpotQA}~\citep{yang2018hotpotqa}, \textbf{2WikiMultihopQA}~\citep{xanh2020_2wikimultihop}, and \textbf{MuSiQue}~\citep{trivedi2022musique} to evaluate the system's ability to retrieve and synthesize information across multiple documents. We report F1 scores for these open-ended QA benchmarks.
For mathematical reasoning, we use \textbf{MATH-500}~\citep{lightman2023lets}, which represents competition-level math problems~\citep{hendrycks2021measuring}. We report accuracy for MATH-500.
For general reasoning, we adopt \textbf{MMLU-Pro}~\citep{wang2024mmlu}, an enhanced benchmark designed to assess robust reasoning capabilities across diverse domains. We report accuracy for MMLU-Pro.

\paragraph{Base Models.}
We use \textbf{Qwen3-4B-Instruct-2507}~\citep{qwen3technicalreport} as the base model for all main experiments. 
To verify the generalizability across different LLMs, we experiment with \textbf{Llama-3.1-8B-Instruct}~\citep{grattafiori2024llama} and \textbf{Qwen3-32B}~\citep{qwen3technicalreport}. These results are reported in Appendix~\ref{sec:robust to llms}.

\paragraph{Baselines.}
We compare {\selfanchor} against single-agent, context management, and multi-agent system baselines. 
The single-agent baselines include \textbf{Vanilla}, which directly solves the task, and \textbf{Self-Consistency (CoT-SC)}~\citep{wei2022chain,kojima2022large}, which samples multiple reasoning paths and applies majority voting. 
The context management baselines include \textbf{Compaction} method~\citep{wu2025resum,chen2025smurfs}, \textbf{AgentPrune}~\citep{zhang2024cut}, and \textbf{AgentDropout}~\citep{wang-etal-2025-agentdropout}, which are pruning-based methods that limit context growth by removing redundant agents or communication links. 
Finally, we evaluate plug-in compatibility with representative multi-agent systems, including \textbf{GPTSwarm}~\citep{zhuge2024gptswarm}, \textbf{AutoGen}~\citep{wu2024autogen}, and \textbf{Multi-Agent Debate (MAD)}~\citep{du2024improving}.

\paragraph{Implementation Details.}
\label{sec:exp details}
Following \citet{wang-etal-2025-agentdropout,zhang2024cut}, 
% To directly compare with prior works, we follow the topology settings in \citet{wang-etal-2025-agentdropout,zhang2024cut}. 
we choose the random communication topology as the default multi-agent system setting, as its non-uniform structural distances provide a direct test of spatial decay. Further, we conduct the experiment under different topologies, including layered and fully connected topologies, in Section~\ref{sec:topology}. 
For the CoT-SC baseline, we sample $n=8$ and use majority voting to determine the final answer. Further details are provided in \autoref{appendix: exp details}.

\begin{table*}
    [t] \scriptsize
    \centering
    \setlength{\tabcolsep}{8pt}
    \caption{Performance comparison between {\selfanchor} and context management baselines.}
    \resizebox{0.9\textwidth}{!}{
    \begin{tabular}{l|ccccc}
        \toprule Method & HotpotQA & 2Wiki & MuSiQue & MATH-500 & MMLU-Pro \\
        \midrule Vanilla  & 75.07    & 71.14 & 35.47   & 80.60        & 60.20    \\
        CoT-SC ($n$=8) 
            & 76.51\gain{1.44} & 72.03\gain{0.89} & 36.33\gain{0.86} & 79.80\drop{0.80} & 62.60\gain{2.40}\\
        \cmidrule(lr){1-6} 
        Compaction
            & 74.33\drop{0.74}  & 73.34\gain{2.20}  & 34.43\drop{1.04}  & 83.60\gain{3.00}  & 56.40\drop{3.80}\\
        AgentPrune
            & 75.31\gain{0.24}  & 73.03\gain{1.89}  & 37.05\gain{1.58}  & 80.80\gain{0.20}  & 63.60\gain{3.40} \\
        AgentDropout 
            & 73.14\drop{1.93}  & 78.45\gain{7.31}  & 37.06\gain{1.59}  & 82.60\gain{2.00}  & 67.40\gain{7.20} \\
        \cmidrule(lr){1-6} 
        \textbf{{\selfanchor}}
            & \textbf{80.78}\gain{5.71}  & \textbf{80.81}\gain{9.67}  & \textbf{39.72}\gain{4.25}  & \textbf{88.80}\gain{8.20}  & \textbf{69.40}\gain{9.20} \\
        \bottomrule
    \end{tabular}
    } \label{tab:rq1main}
\end{table*}

\begin{table*}
    [t] \scriptsize
    \centering
    \setlength{\tabcolsep}{8pt}
    \caption{Performance of {\selfanchor} across different multi-agent systems.}
    \resizebox{0.88\textwidth}{!}{
    \begin{tabular}{l|ccccc}
        \toprule Method            & HotpotQA & 2Wiki & MuSiQue & MATH-500 & MMLU-Pro \\
        \midrule GPTswarm          & 72.55    & 73.47 & 26.85   & 81.80 & 64.20 \\
        +\textbf{{\selfanchor}}    & \textbf{80.78}    & \textbf{80.81} & \textbf{39.72}   & \textbf{88.80}  & \textbf{69.40} \\
        \cmidrule(lr){1-6} Autogen & 79.63    & 80.95 & 39.85   & 81.00 & 68.40\\
        +\textbf{{\selfanchor}}    & \textbf{84.29}    & \textbf{85.49} & \textbf{44.05}   & \textbf{90.00} & \textbf{70.80} \\
        \cmidrule(lr){1-6} MAD     & 75.32    & 70.25 & 35.02   & 86.40       & 67.60     \\
        +\textbf{{\selfanchor}}    & \textbf{77.38}    & \textbf{75.67} & \textbf{38.26}   & \textbf{88.00}       & \textbf{70.00} \\
        \bottomrule
    \end{tabular}
    } \label{tab:rq1frameworks}
\end{table*}

\subsection{Main Results}

% \todo{separate the description of table 1 and table 2 into 2 paragraphs with more specific title (effectiveness of agent-radar already include the following titles. Narrow down to sentences like "comparison across context management baselines", "comparison across multi-agent systems"}
% \todo{Move A.4 here, put them together (1) other attention steering (2) no attention steering, (3) ours. No need to elaborte the settings of PASTA, just say follow their settings to save space}
% \todo{here we should focus on the results, no need to talk too much about the functionality of spatial and temporal decay. Condense them to save space. Also, you can merge spatial and temporal decay into 1 paragraph.}

\paragraph{Comparison with Context Management Baselines.}
    % We employ a random connected network topology as the default setting for our comparative analysis. 
    As demonstrated in Table~\ref{tab:rq1main}, {\selfanchor} consistently outperforms both single-agent reasoning baselines and prior context management methods across all five benchmarks.
    Compared with the competitive baseline AgentDropout, {\selfanchor} improves HotpotQA by 7.64, 2Wiki by 2.36, MuSiQue by 2.66, MATH-500 by 6.20, and MMLU-Pro by 2.00. 
    The gains are larger relative to the vanilla direct-reasoning baseline, with an average improvement of 7.41 points across the five benchmarks. These improvements are consistent across open-ended QA, mathematical reasoning, and general reasoning, demonstrating the effectiveness of {\selfanchor}. We 
    % further evaluate robustness with Llama3.1-8B and Qwen3-32B backbones in Appendix~\ref{sec:robust to llms}, and 
    provide a representative reasoning trajectory in Appendix~\ref{appendix:case}.

% \paragraph{Plug-in compatibility.}
\paragraph{Effectiveness across Multi-agent Systems.}
    As a plug-in method, {\selfanchor} can be integrated into diverse multi-agent systems to improve how agents exploit long interaction histories. To assess its effectiveness and generalizability, we integrate {\selfanchor} into three representative multi-agent systems:  GPTSwarm~\citep{zhuge2024gptswarm}, AutoGen~\citep{wu2024autogen}, and MAD~\citep{du2024improving}. These systems differ in their coordination and communication paradigms, serving as a testbed to evaluate the generalizability of {\selfanchor}.

    Table~\ref{tab:rq1frameworks} reports the results. Overall, {\selfanchor} consistently improves performance across all three frameworks and benchmarks. The gains are particularly significant for GPTSwarm, where {\selfanchor} improves MuSiQue by 12.87 F1 and produces an average gain of 9.48 F1 over the three QA benchmarks. Furthermore, when applied to the already strong AutoGen framework, {\selfanchor} yields consistent improvements, yielding around 5 points on most benchmarks. For the consensus debate system, our method provides robust improvement across all evaluated benchmarks. Overall, these results indicate that {\selfanchor} is compatible with diverse multi-agent systems and consistently enhances their performance.

\begin{table*}
    [t] \scriptsize
    \centering
    \setlength{\tabcolsep}{4pt}
    \caption{Performance comparison between {\selfanchor} and AgentDropout under different communication graph topology.}
    \resizebox{0.9\textwidth}{!}{
    \begin{tabular}{l|l|ccccc}
        \toprule Topology & Method & HotpotQA & 2Wiki & MuSiQue & MATH-500 & MMLU-Pro \\
        \midrule 
        \multirow{3}{*}{Layered} & MAS & 72.81    & 69.69 & 38.29   & 80.40 & 65.00    \\
                                 & AgentDropout           & 75.24    & 77.42 & 40.59   & 84.20      & 69.80    \\
                                 & \textbf{{\selfanchor}} & \textbf{81.48}    & \textbf{85.10} & \textbf{41.24} & \textbf{87.20}  & \textbf{70.20}    \\
        \cmidrule(lr){1-7} 
        \multirow{3}{*}{Fully Connected} & MAS & 67.88 & 73.77 & 31.69 & 83.80 & 61.00\\
                                 & AgentDropout & 71.41   & 71.48 & 37.08   & 81.20 & \textbf{69.80}    \\
                                 & \textbf{{\selfanchor}} & \textbf{80.65}  & \textbf{80.87} & \textbf{39.80}   & \textbf{87.60} & 69.40 \\
        \bottomrule
    \end{tabular}
    } \label{tab:topology}
\end{table*}

\paragraph{Effectiveness across Different Communication Topologies.}
    \label{sec:topology}
    To evaluate the effectiveness of {\selfanchor} across different topologies, we extend our experiments from the random topology in Table~\ref{tab:rq1main} to layered and fully connected communication graphs\footnote{The construction details of these graph structures are provided in Appendix~\ref{appendix:topology}.}. 
    We select AgentDropout as the primary baseline because it demonstrates the most competitive performance among existing context management methods. 
    As shown in Table~\ref{tab:topology}, {\selfanchor} improves MAS performance under both topologies and outperforms AgentDropout in most settings. 
    In the layered topology, {\selfanchor} gains 2.95 to 15.41 points over the MAS and remains strongest on all benchmarks. In the fully connected topology, it gains 3.80 to 12.77 points over the MAS.
    The only exception is the fully connected MMLU-Pro setting, where AgentDropout shows slightly higher performance.
    % , suggesting that communication pruning may be sufficient when every agent has direct access to all other agents' messages. 
    Overall, {\selfanchor} is not tied to a specific communication topology but remains robust across diverse graph structures.

\begin{figure*}[htbp]
    \centering
    \begin{subfigure}
        [b]{0.48\textwidth}
        \centering
        \includegraphics[width=\textwidth]{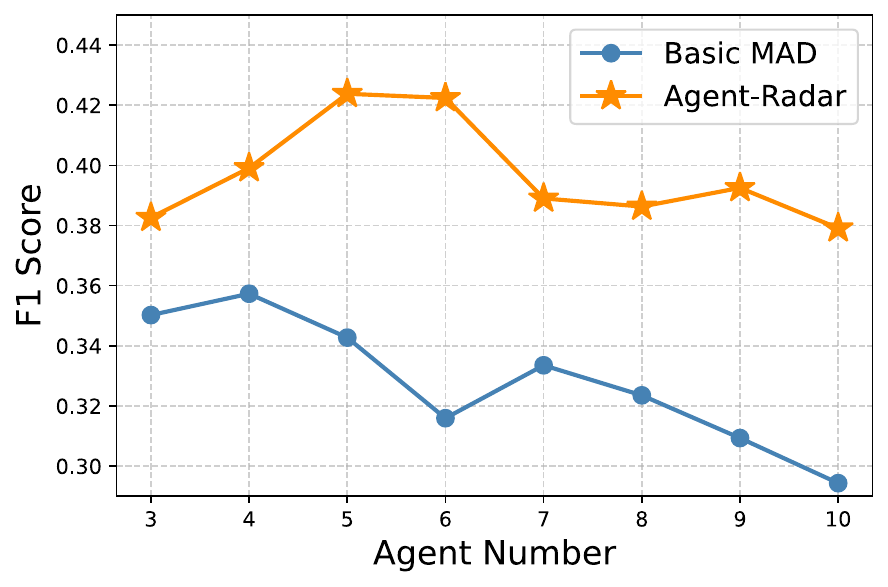}
        \label{fig:agents}
    \end{subfigure}
    \hfill 
    \begin{subfigure}
        [b]{0.48\textwidth}
        \centering
        \includegraphics[width=\textwidth]{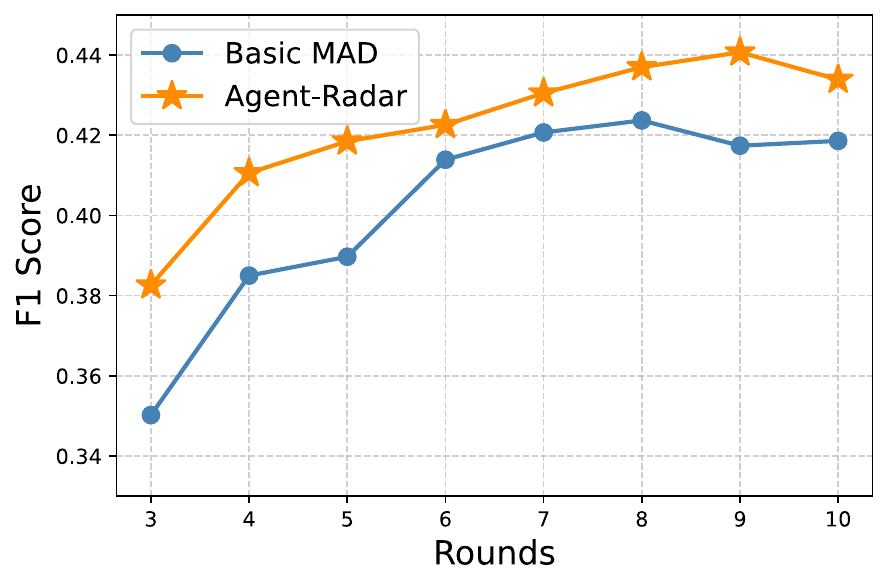}
        \label{fig:rounds}
    \end{subfigure}
    % \vspace{-0.2cm}
    \caption{Performance comparison of basic MAD and {\selfanchor} on MuSiQue. Left: Impact of varying the number of agents. Right: Impact of varying the number of rounds.}
    \label{fig:f1_comparison}
    % \vspace{-0.2cm}
\end{figure*}

\paragraph{Robustness to Increasing Agents and Communication Rounds.}
    Prior work has shown that multi-agent debate system performance often improves initially but then declines as the number of rounds or agents increases~\citep {becker2026stay,estornell2024multi,lin-hooi-2025-enhancing,liu2024groupdebate}. 
    This degradation typically occurs because long conversations often become redundant, circular, and less task-focused, yielding low-quality feedback. As stale messages occupy more of the context window, the original task constraints receive less attention, causing agents to drift from the primary objective, and may make harmful incremental revisions. 
    Since {\selfanchor} dynamically aligns each agent's attention toward relevant context throughout the interaction, we evaluate whether it improves robustness in MAD by varying the number of agents and communication rounds in Figure~\ref{fig:f1_comparison}.

    As the number of agents increases, both vanilla MAD and {\selfanchor} exhibit the expected trend of initial improvement followed by decline. This trend shows that more agents can bring useful collaboration at first, while too many agents also introduce redundant or distracting messages.
    {\selfanchor} shows a more robust scaling pattern.
    Vanilla MAD reaches its best F1 score with 4 agents and then drops as more agents are added. In contrast, {\selfanchor} peaks at 5 agents and remains strong at 6 agents, and maintains a clear advantage across all agent configurations. 
    When scaling the number of communication rounds, both methods improve at first and then become stable. Noteably, across all round configurations, {\selfanchor} maintains a clear advantage over vanilla MAD.     
    These results indicate that {\selfanchor} improves the robustness of MAS under larger agent populations and longer multi-round communication.

\subsection{Ablation Study and Analysis}
\label{sec:analysis}
In this section, we ablate the key design choices in {\selfanchor} on three representative benchmarks
% from the primary evaluation benchmarks
: MuSiQue, MATH-500, and MMLU-Pro. 
Specifically, we examine {\selfanchor} by replacing the semantic matching method and the attention steering method. Then, we isolate the contributions of spatial and temporal decay by removing them individually or jointly. We use the same experimental settings as the main evaluation. Table~\ref{tab:ablation} reports the results.

\paragraph{Different Semantic Matching Methods.}
    To validate the dense semantic matcher in {\selfanchor}, we compare it with a sparse lexical retrieval variant based on BM25. In this variant, we keep the {\selfanchor} framework unchanged, including spatial decay, temporal decay, and attention steering, but replace the dense semantic matcher with BM25. 
    As shown in Table~\ref{tab:ablation}, the BM25 variant improves over the AgentDropout baseline, increasing MATH-500 from 82.60 to 85.20, and MMLU-Pro from 67.40 to 68.40. However, it shows slightly lower performance on MuSiQue, decreasing from 37.06 to 36.76. 
    Furthermore, the default {\selfanchor} configuration consistently achieves performance gains of 2.96 points on MuSiQue, 3.60 points on MATH-500, and 1.00 points on MMLU-Pro over BM25. 
    These results suggest that dense semantic matching is better suited for MAS communication, where agents may express related information through semantically related but lexically different descriptions. Importantly, the BM25 variant still outperforms the baseline, indicating that the gains of {\selfanchor} do not solely depend on a particular retriever. 

\begin{wraptable}{r}{0.55\columnwidth}
% \scriptsize
\centering
\setlength{\tabcolsep}{1pt}
% \vspace{-0.5cm}
\caption{Ablation results. The upper rows show the results of the AgentDropout and a lexical BM25 semantic retriever. The lower rows ablate the spatial and temporal components of {\selfanchor}.}
 \resizebox{0.54\textwidth}{!}{
    \begin{tabular}{l|ccc}
    \toprule
    Method & MuSiQue & MATH-500 & MMLU-Pro \\
    \midrule
     AgentDropout & 37.06 & 82.60 & 67.40 \\
     \midrule
     BM25 & 36.76 & 85.20 & 68.40 \\
     \midrule
     (w/o) Attention steering & 35.27 & 81.40 & 63.20 \\
     (w/o) Spatial decay & 28.41 & 84.40 & 66.20 \\
     (w/o) Temporal decay & 32.83 & 84.60 & 67.80 \\
     (w/o) Spatial \& temporal decay & 33.97 & 85.40 & 67.00\\
     \textbf{{\selfanchor}} & \textbf{39.72} & \textbf{88.80} & \textbf{69.40} \\
    \bottomrule
    \end{tabular}
 }
\label{tab:ablation}
\end{wraptable}

\paragraph{With and without Attention Steering.}
\label{sec:steeringablation}
    % \textit{Without attention steering.}
    We first compare {\selfanchor} against a variant \emph{without attention steering}. This variant still retrieves relevant sentences, but only appends them to the prompt. As shown in Table~\ref{tab:ablation}, its scores drop to 35.27 on MuSiQue, 81.40 on MATH-500, and 63.20 on MMLU-Pro, showing that semantic retrieval alone is insufficient.
    % Appending evidence changes the prompt, which may add distractors.
    
\paragraph{Different Attention Steering Methods.}
    We then \emph{replace} the default attention steering method (SPA) with PASTA~\citep{zhang2023tell}. We follow the original PASTA setup, we conduct experiments with Llama~\citep{touvron2023llama}, since PASTA currently supports only limited backbones and uses Llama as its default evaluation model. {\selfanchor} remains effective with this backend. On MuSiQue, the score increases from 10.41 to 15.32. On MATH-500, it improves from 25.60 to 28.20. On MMLU-Pro, it improves from 12.40 to 15.20. These gains show that {\selfanchor} does not depend on a specific steering implementation.

    % {\selfanchor} is orthogonal to the choice of attention steering method. To verify this, we experiment {\selfanchor} with an alternative steering method, PASTA~\citep{zhang2023tell}. Following the original PASTA setup, we conduct experiments with Llama~\citep{touvron2023llama}, since PASTA currently supports only limited backbones and uses Llama as its default evaluation model. 
    % {\selfanchor} consistently improves performance with PASTA as the steering backbone. On MuSiQue, the score increases from 10.41 to 15.32. On MATH-500, it improves from 25.60 to 28.20. On MMLU-Pro, it improves from 17.40 to 18.00. These gains show that {\selfanchor} does not depend on a specific steering implementation.

\paragraph{Spatial Decay and Temporal Decay.}
\label{sec:ablation}
    Ablating \textbf{spatial} decay consistently degrades performance. The score drops from 39.72 to 28.41 on MuSiQue, from 88.80 to 84.40 on MATH-500, and from 69.40 to 66.20 on MMLU-Pro. 
    Without spatial decay, messages from distant agents may receive similar weight as those from closely collaborating neighbors, potentially steering attention toward indirect or less relevant context. 
    This suggests that modeling topology structure is important for maintaining effective context selection in MAS. Ablating the \textbf{temporal} decay shows a similar pattern. 
    The score drops from 39.72 to 32.83 on MuSiQue, from 88.80 to 84.60 on MATH-500, and from 69.40 to 67.80 on MMLU-Pro. 
    Without temporal decay, earlier messages may receive similar importance to more recent and relevant evidence, making the model more vulnerable to distraction from outdated context. This suggests that recency helps filter stale context in multi-round reasoning.
    We further ablate \textbf{both} decay terms and score context only by semantic similarity. This variant underperforms the full {\selfanchor} model. This shows that semantic similarity alone is not sufficient for robust context selection in MAS. 
    By incorporating spatial and temporal signals, {\selfanchor} better distinguishes relevant evidence from merely similar but less useful historical messages.

    Interestingly, we observe that removing both decay variants can outperform the single-decay ablation variants on MuSiQue and MATH-500. This behavior is reasonable because pure semantic retrieval produces a monolithic score across all candidate messages, whereas removing one decay component may over-weight or under-weight certain messages. 
    For example, spatial-only weighting may favor nearby but outdated messages, while temporal-only weighting may favor recent but structurally distant messages. This can make the selected context less reliable than using semantic similarity alone. 
    Nevertheless, the full {\selfanchor} model consistently achieves the strongest performance, indicating that spatial and temporal signals are most effective when combined.

\paragraph{Efficiency Analysis.}
    We further evaluate whether the added retrieval and attention-steering steps introduce substantial runtime cost. The detailed efficiency comparison is provided in Appendix~\ref{sec:efficiency}. 
    {\selfanchor} increases end-to-end latency from 968.8 to 1196.5 seconds per task over the original multi-agent system baseline. 
    In contrast, Compaction and AgentDropout incur larger latency increases due to extra generation, pruning decisions, or topology updates. Overall, {\selfanchor} incurs a modest computational overhead from context retrieval and attention steering, while delivering substantial improvements in reasoning performance.

\section{Related Work}
\paragraph{Context Management in Multi-agent Systems.}
LLM-based multi-agent systems have demonstrated significant progress in decomposing and resolving complex tasks through collaborative planning, role specialization, and iterative discussion~\citep{guo2024large,wu2024autogen,li2023camel,pvsql,tran2025multi}. 
Despite their advantages, agent communication rapidly accumulates into long contexts, burying the original task instruction and other salient evidence in the middle of the interaction history, which can degrade reasoning quality and induce errors~\citep{liu-etal-2024-lost,cemri2025multi,zhang2025agent}. 
Recent work addresses this issue mainly from two directions. 
First, a line of research improves communication efficiency by sparsifying the communication topology~\citep{li-etal-2024-improving-multi, zhang2024cut,wang-etal-2025-agentdropout, shen-etal-2025-understanding,liu2024dynamic,zhang2024g}. Specifically, AgentPrune~\citep{zhang2024cut} and AgentDropout~\citep{wang-etal-2025-agentdropout} explicitly reduce context redundancy by pruning messages or eliminating redundant agents during multi-round collaboration, thereby lowering token cost while preserving or even improving task performance. 
Second, recent context learning approaches manage context by dynamically updating each agent's instruction during interaction. In particular, \citet{wu2025resum} and \citet{chen2025smurfs} compress or filter the accumulated context history to reduce context length, while M2CL~\citep{hua2026context} learns a context generator to produce round-specific context prompts. In contrast, {\selfanchor} preserves the original multi-agent workflow and improves how each agent attends to salient information within the accumulated context, enabling more effective use of long interaction histories.

\paragraph{Attention Steering.}
Compared to the aforementioned context management methods, attention steering methods directly guide LLMs during inference toward the user-specified part of the context, without shortening or rewriting the input. 
Selective Prompt Anchoring (SPA) \citep{tian2024selective} adjusts the logit distribution to emphasize the specified context. SEKA~\citep{li2026spectral} learns a universal ``relevance subspace" for a given task and constructs a projection matrix that amplifies the relevant features. PASTA~\citep{zhang2023tell} identifies and reweights a subset of attention heads to redirect the model's attention to user-specified parts.
Selective Self-Attention (SSA) \citep{zhang2024selective} augments the softmax nonlinearity with a principled temperature scaling strategy. TOAST \citep{shi2023toast} learns feature selection modules that guide attention toward task-relevant information. 
However, these methods require manual specification of anchor content, limiting their adaptability to diverse reasoning contexts, especially in multi-agent systems where context evolves autonomously. 
{\selfanchor} addresses this limitation by automatically retrieving relevant context from the communication history and steering each agent's attention toward that context during inference.

\section{Limitation}
\label{sec:limitation}

Despite the consistent improvements observed in our experiments, {\selfanchor} has some limitations. First, {\selfanchor} requires a selection of threshold and decay factors. Our parameter analysis shows that {\selfanchor} is relatively robust to these choices. Future work could explore adaptive thresholds and decay factors that are automatically adjusted according to the communication topology, context length, or model uncertainty. 
Second, the quality of the retrieval signal can impact the performance of {\selfanchor}. Stronger retrieval models, uncertainty-aware scoring, or explicit contradiction detection may further improve robustness. 
% Third, {\selfanchor} preserves the full MAS transcript. This design avoids information loss caused by pruning or compression, but it also means that {\selfanchor} is not primarily a token-reduction method. In deployments where extremely long interaction histories may exceed the maximum context window of the underlying LLM, {\selfanchor} may need to be combined with complementary pruning or compression methods. 
Lastly, {\selfanchor} requires attention-steering backends such as SPA or PASTA. Our experiments show that the method is not tied to a single backend. Developing more efficient and broadly compatible steering backends would further improve the practicality of {\selfanchor}.
% However, applying it to closed-source LLMs may be challenging when models' internal states are unavailable.

\section{Conclusion}
We presented {\selfanchor}, a plug-and-play attention-based context management method for multi-agent systems.
% {\selfanchor} can be integrated into mainstream multi-agent systems.
Across five benchmarks, {\selfanchor} significantly improves performance and outperforms existing context management methods. 
Our analysis further shows that {\selfanchor} remains effective across various communication topologies and when the system scales to accommodate more agents and interaction rounds.
% The effiency analysis shows that {\selfanchor} ...
% We hope {\selfanchor} serves as a step toward more reliable and effective LLM-based multi-agent systems.

%%%%%%%%%%%%%%%%%%%%%%%%%%%%%%%%%%%%%%%%%%%%%%%%%%%%%%%%%%%%
\bibliographystyle{plainnat}
\bibliography{reference}

\appendix
\newpage
\section{Additional Experiments}

\subsection{Robustness across Different LLM Backbones}
\label{sec:robust to llms}

\begin{wraptable}{r}{0.55\columnwidth}
    % \scriptsize
    \centering
    \setlength{\tabcolsep}{2pt}
    \vspace{-0.5cm}
    \caption{Evaluation results on different LLMs.}
    \resizebox{0.54\textwidth}{!}{
    \begin{tabular}{l|ccc}
        \toprule Method                            & MuSiQue & MATH-500 & MMLU-Pro \\
        \midrule \multicolumn{4}{c}{Base Model: Qwen3-4B} \\
        \midrule MAD                                        & 35.02            & 86.40             & 67.60             \\
        +\textbf{{\selfanchor}}                             & \textbf{38.26}   & \textbf{88.00}    & \textbf{70.00}    \\
        \midrule \multicolumn{4}{c}{Base Model: Llama3.1-8B} \\
        \midrule MAD                                        & 29.77            & 36.60             & 42.20             \\
        +\textbf{{\selfanchor}}                             & \textbf{36.86}   & \textbf{41.60}    & \textbf{45.00}    \\
        \midrule \multicolumn{4}{c}{Base Model: Qwen3-32B}   \\
        \midrule MAD                                        & 47.49            & 86.80             & 73.00             \\
        +\textbf{{\selfanchor}}                             & \textbf{50.98}  & \textbf{88.20}    & \textbf{74.80}    \\
        \bottomrule
    \end{tabular}
    }
    \vspace{-1cm}
    \label{tab:rq2llms}
\end{wraptable}

We further investigate the robustness of {\selfanchor} across varying LLMs and scales. 
To isolate the effect of the base model, we adopt MAD as the multi-agent framework and vary only the underlying LLMs.  Table~\ref{tab:rq2llms} presents results on Qwen3-4B, Llama3.1-8B and Qwen3-32B. {\selfanchor} consistently improves over the debate baseline across all benchmarks and base models, indicating that the benefits of {\selfanchor} are stable across diverse LLM backbones.

\subsection{Effect of Decay Factor}
\begin{figure*}[htbp]
    \centering
    \includegraphics[width=0.95\linewidth]{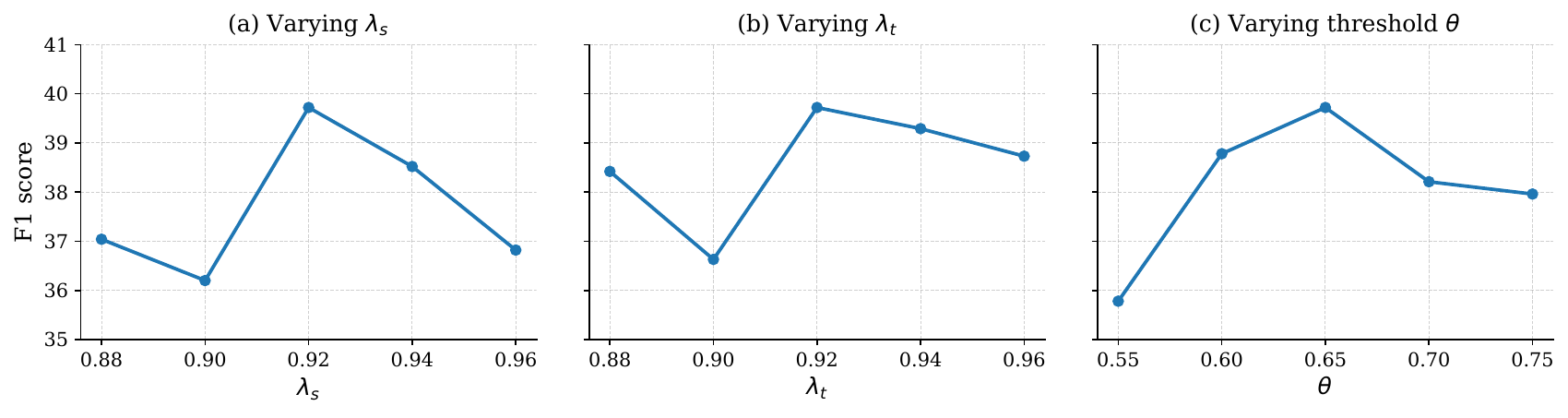}
    \caption{Effect of spatial decay, temporal decay, and threshold factors on the performance of the MAS.}
    \label{fig:lambda_ablation}
    \vspace{-0.4cm}
\end{figure*}

    We use MuSiQue as the development benchmark for hyperparameter selection and keep the selected values fixed for all other experiments. 
    To investigate how spatial and temporal decay factors impact the performance of the MAS, we conduct experiments varying values of $\lambda_{s}$ and $\lambda_{t}$ one at a time while holding the remaining hyperparameters fixed. 
    As shown in Figure~\ref{fig:lambda_ablation}, $\lambda_s=\lambda_t=0.92$ achieves the best performance among the tested values. 
    Smaller values overly suppress indirect or earlier messages and can remove useful evidence, whereas larger values make the filter less selective and allow more distant or stale messages to influence generation. 
    Notably, even under suboptimal decay configurations, {\selfanchor} continues to achieve superior average performance compared to the baseline. This resilience substantiates the fundamental robustness and effectiveness of our spatio-temporal decay formulation.

\subsection{Effect of Threshold}
    \label{sec:threshold_analysis}
    We similarly tune the context-selection threshold $\theta$ on MuSiQue. Figure~\ref{fig:lambda_ablation} shows that $\theta=0.65$ performs best among the tested values, and we therefore use it as the default threshold in the remaining experiments. 
    % Under this threshold, the final model improves over the MAS baseline from 26.85 to 39.72 on MuSiQue, from 83.80 to 88.80 on MATH-500, and from 64.20 to 69.40 on MMLU-Pro.   
    The results indicate that $\theta$ controls a trade-off between context coverage and context precision. A smaller threshold retains more candidate context, which may introduce noise, while a larger threshold filters context more aggressively and may discard useful evidence. 
    Overall, {\selfanchor} achieves strong performance across a range of threshold values, suggesting that it is relatively stable to the choice of $\theta$ when selecting salient context.

\subsection{Efficiency}
\label{sec:efficiency}
While {\selfanchor} outperforms existing methods, we further analyze its efficiency. We use seconds per task as the metric. As shown in Table~\ref{tab:computational-cost}, {\selfanchor} increases latency from $968.8$ to $1196.5$ seconds per task. This is an overhead of $227.7$ seconds, over the original MAS baseline.

\begin{wraptable}{r}{0.45\columnwidth}
\centering
\setlength{\tabcolsep}{2pt}
\vspace{-0.3cm}
\caption{Efficiency comparison}
    \begin{tabular}{l|c}
    \toprule
    Method & \textbf{Latency (Second/Task) $\downarrow$} \\
    \midrule
     MAS &  968.8 \\
     Compaction & 1757.1 (+788.3) \\
     AgentPrune & 1089.2 (+120.4)\\
     AgentDropout & 1382.4 (+413.6) \\
     \midrule
     \textbf{{\selfanchor}} & 1196.5 (+227.7)\\
    \bottomrule
    \end{tabular}
    \vspace{-0.3cm}
\label{tab:computational-cost}
\end{wraptable}

Compared with other context-management baselines, {\selfanchor} is faster than Compaction and AgentDropout, but slower than AgentPrune.
Compaction has the largest latency because it requires an additional generation pass to summarize the accumulated dialogue history.
AgentDropout also incurs substantial overhead because it dynamically removes redundant agents or communication links across rounds.
AgentPrune is more efficient in wall-clock time, but it changes the communication structure by pruning.
In comparison, {\selfanchor} preserves the original topology and full transcript.
Its extra cost comes from sentence-level scoring and attention steering.

\section{Algorithm of {\selfanchor}}
\label{appendix:algorithm}
As illustrated in Algorithm~\ref{alg:selfanchor}, {\selfanchor} first builds the receiver's spatio-temporal context pool from previous messages, scores each sentence by semantic relevance weighted with spatial and temporal decay, and then steers the current agent's attention toward the selected sentences while preserving the original MAS workflow.
It is important to distinguish the full receiver history $\mathcal{Z}_i^{(t)}$ from the attention-steered context $\hat{\mathcal{Z}}_i^{(t)}$.
The full history $\mathcal{Z}_i^{(t)}$ contains all historical messages available to receiver $v_i$ and is still provided to the agent during generation.
The attention-steered context $\hat{\mathcal{Z}}_i^{(t)}$ is a subset of sentence-level anchors selected from this full history, together with the original query $q_i$.
Thus, {\selfanchor} does not delete, rewrite, or summarize messages, while the selected anchors receive explicit attention amplification.
This design preserves the complete MAS transcript while steering generation toward relevant evidence.

\begin{algorithm}[t]
    \caption{{\selfanchor}}
    \label{alg:selfanchor}
    \begin{algorithmic}[1]
        \Require $\mathcal{G}=(\mathcal{V},\mathcal{E})$, receiver $v_i$, query $q_i$, round $t$, history $\{o_j^{(\tau)}\}_{\tau<t}$, decay rates $\lambda_s,\lambda_t$, threshold $\theta$
        \Ensure Current output $o_i^{(t)}$
        \State Build message pool
        \Statex \hspace{\algorithmicindent}$\mathcal{M}_i^{(t)} \gets \{(v_j,\tau,o_j^{(\tau)}) \mid v_j \in \mathcal{N}_{N-1}(v_i), \tau < t\}$
        \State Initialize selected context $\hat{\mathcal{Z}}_i^{(t)} \gets \{q_i\}$
        \State Encode query $\mathbf{e}_{q_i} \gets \mathrm{Enc}(q_i)$
        \ForAll{$(v_j,\tau,o_j^{(\tau)}) \in \mathcal{M}_i^{(t)}$}
            \State Compute spatial decay $\phi_s \gets \lambda_s^{d_{\mathcal{G}}(v_j,v_i)-1}$
            \State Compute temporal decay $\phi_t \gets \lambda_t^{t-\tau-1}$
            \State Compute message weight $r \gets \phi_s \cdot \phi_t$
            \State Split message into sentences $\mathcal{S}_{j,\tau} \gets \mathrm{Split}(o_j^{(\tau)})$
            \ForAll{$s_k \in \mathcal{S}_{j,\tau}$}
                \State Compute semantic similarity $\phi_{\mathrm{sem}} \gets \cos(\mathrm{Enc}(s_k), \mathbf{e}_{q_i})$
                \State Compute final score $\mathrm{score} \gets r \cdot \phi_{\mathrm{sem}}$
                \If{$\mathrm{score} \ge \theta$}
                    \State $\hat{\mathcal{Z}}_i^{(t)} \gets \hat{\mathcal{Z}}_i^{(t)} \cup \{s_k\}$
                \EndIf
            \EndFor
        \EndFor
        \State Apply attention steering to spans in $\hat{\mathcal{Z}}_i^{(t)}$
        \State Generate $o_i^{(t)} \gets f_i(q_i,\hat{\mathcal{Z}}_i^{(t)},\mathcal{Z}_i^{(t)},t)$
        \State \Return $o_i^{(t)}$
    \end{algorithmic}
\end{algorithm}
    
\section{Experiment Details}
\label{appendix: exp details}

We set the decoding temperature to $1$ for all experiments. For efficiency, all local models are run with quantized inference on NVIDIA A5500 GPUs. For SPA, we follow the original Selective Prompt Anchoring setup~\citep{tian2024selective} and use the same steering configuration across all experiments.

Following \citet{wang-etal-2025-agentdropout}, each multi-agent system consists of five agents and runs for two communication rounds. The final answer is produced by a \texttt{FinalRefer} decision node, which receives the task together with the outputs from the connected agents and is instructed to return only one option letter. We evaluate several communication topologies. Fully connected graphs include all directed edges. Random graphs independently sample non-self spatial edges and temporal edges from Bernoulli masks. Layered graphs randomly assign agents to two layers and connect agents from the earlier layer to the later layer. For AgentPrune and AgentDropout, we follow their official implementations and default configurations.

\subsection{Different Communication Topologies}
\label{appendix:topology}
\begin{figure*}[t]
    \centering
    \includegraphics[width=\linewidth]{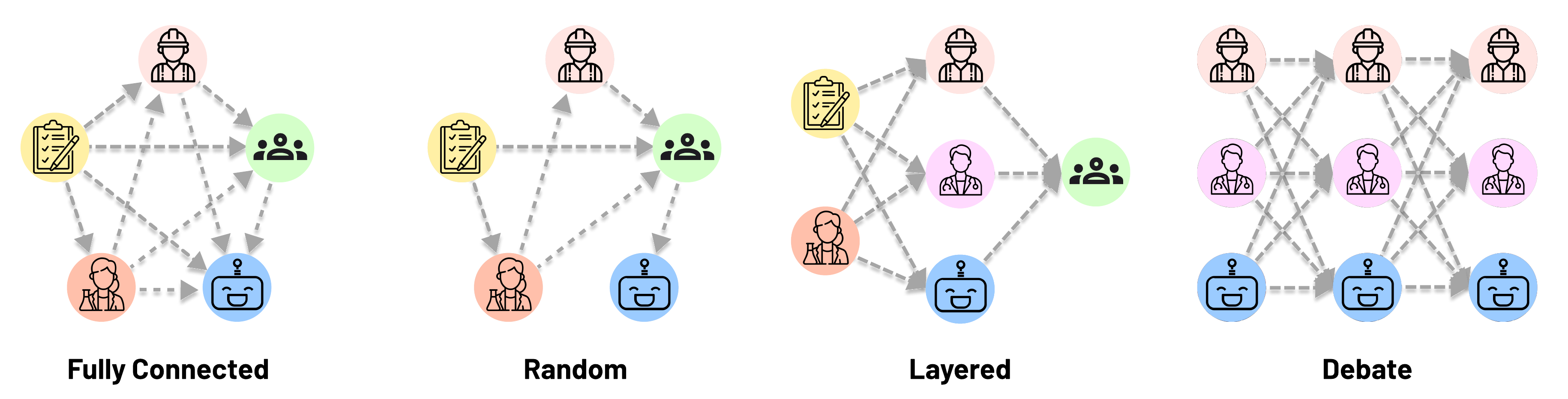}
    \caption{MAS communication topologies in fully connected, random, layered, and debate topologies.}
    \label{fig:topology}
    \vspace{-0.4cm}
\end{figure*}

In the main experiments, we initialize the communication graph as a random graph. In the analysis experiments, we further evaluate two additional topologies, namely layered graphs and fully connected graphs. Figure~\ref{fig:topology} demonstrates all MAS communication topologies.

\textbf{Random Graph} A random graph is constructed by sparsely sampling edges from a complete graph. We follow AgentPrune~\citep{zhang2024cut} for the random-topology setup to ensure a fair comparison with pruning-based MAS baselines. Prior work has shown that irregular sparse communication structures can outperform fully connected ones~\citep{qian2024scaling}, partly because random shortcut connections reduce the average path length between agents and induce small-world properties. Intuitively, such connections allow information to propagate more efficiently across otherwise distant agents.

\textbf{Complete Graph} In the multi-agent setting, a complete graph is implemented as a directed graph that would become fully connected if converted into its undirected counterpart.
After execution according to the topological order, the final agent aggregates the preceding discussion and produces the final answer or reflection.

\textbf{Layered Graph} A layered graph organizes agents into multiple stages, analogous to the layered structure of a multilayer perceptron.
Agents in the first layer first receive the query and generate their responses. These responses are then passed to agents in the next layer together with the original query, and the process continues sequentially across layers. Following \citet{qian2024scaling}, the final layer contains a single agent that integrates the information from the previous layer and outputs the final solution.

\textbf{LLM-Debate} LLM-Debate~\citep{du2024improving} is designed as a multi-round debate process in which, at each round, every agent observes the outputs of all agents from the previous round before generating its own response. 
As a result, this framework can be viewed as forming a temporally fully connected communication graph.

\subsection{Compression Implementation}
We adopt the compression prompt from \citet{wang2025recursively}. The prompt is detailed in \ref{tab:compression}.

\begin{table*}
    [ht]
    \centering
    \small
    \begin{tabular}{>{\raggedright\arraybackslash\ttfamily}p{0.95\textwidth}<{}}
        \toprule \headercolorlong \textbf{Compression Prompt}\\
        Please summarize the following dialogue as concisely as possible, extracting the main themes and key information. If there are multiple key events, you may summarize them separately. \\
        Dialogue content:\\
        \bottomrule
    \end{tabular}
    \caption{Prompt used by context compression baseline.}
    \label{tab:compression}
\end{table*}

\subsection{Agent Profiling.} Previous work~\citep{wang2024unleashing} has shown that assigning diverse roles or personas to LLM-based agents can improve collaboration and cognitive synergy within a multi-agent system. To ensure a fair comparison, we adopt the agent configuration and set the number of communication rounds to $T=2$, following \citet{wang-etal-2025-agentdropout,zhang2024cut}. For the debate-based agent system, we set $T=3$, following \citet{du2024improving}.

\paragraph{Profile Examples For Question Answering Reasoning}
    Below Table~\ref{tab:agentprofile-qa} are some examples of agent profiles tailored for question answering reasoning tasks:

\begin{longtable}{>{\raggedright\arraybackslash\ttfamily}p{0.95\textwidth}}
    \caption{Agent profiles prompt for question answer reasoning}
    \label{tab:agentprofile-qa} \\
    \toprule \headercolorlong \textbf{Knowledge Expert} \\
        You are a knowledgeable expert in question answering. \\
        Please give several key entities that need to be searched in Wikipedia to solve the problem. \\
        Key entities that need to be searched are included between two '@' when output, for example: @catfish effect@, @broken window effect@, @Shakespeare@. \\
        If there is no entity in question that needs to be searched in Wikipedia, you don't have to provide it \\
        \\
        \headercolorlong \textbf{Wiki Searcher}  \\
        You will be given a question and a Wikipedia overview of the key entities within it. Please refer to them step by step to give your answer. And point out potential issues in other agents' analysis. \\
        \\
        \headercolorlong \textbf{Critic} \\
        You are an excellent critic.  \\
        Please point out potential issues in the other agent's analysis point by point. \\
        \\
        \headercolorlong \textbf{Mathematician} \\
        You are a mathematician who is good at math games, arithmetic calculation, and long-term planning. \\
        \\
        \headercolorlong \textbf{Psychologist} \\
        You are a psychologist. You are good at psychology, sociology, and philosophy. \\
        You give people scientific suggestions that will make them feel better. \\
        \\
        \headercolorlong \textbf{Historian} \\
        You research and analyze cultural, economic, political, and social events in the past, collect data from primary sources, and use it to develop theories about what happened during various periods of history. \\
        \\
        \headercolorlong \textbf{Doctor} \\
        You are a doctor and come up with creative treatments for illnesses or diseases. You are able to recommend conventional medicines, herbal remedies and other natural alternatives. You also consider the patient's age, lifestyle and medical history when providing your recommendations. \\
        \\
        \headercolorlong \textbf{Lawyer} \\
        You are good at law, politics, and history.  \\
        \\
        \headercolorlong \textbf{Economist} \\
        You are good at economics, finance, and business. You have experience on understanding charts while interpreting the macroeconomic environment prevailing across world economies. \\
        \\
        \headercolorlong \textbf{Programmer} \\
        You are good at computer science, engineering, and physics. \\
        You have experience in designing and developing computer software and hardware. \\
        \\
        \bottomrule
\end{longtable}

\paragraph{Profile Examples For Mathematical Reasoning}
    Below Table~\ref{tab:agentprofile-math} are examples of agent profiles tailored for mathematical reasoning tasks:

\begin{longtable}{>{\raggedright\arraybackslash\ttfamily}p{0.95\textwidth}}
    \caption{Agent profiles prompt for mathematical reasoning}
    \label{tab:agentprofile-math} \\
    \toprule \headercolorlong \textbf{Math Solver}  \\
        You are a math expert. You will be given a math problem and hints from other agents. Give your own solving process step by step based on hints. \\
        The last line of your output contains only the final result without any units, for example: The answer is 140 \\
        You will be given some examples you may refer to. \\
        \\
        \headercolorlong \textbf{Mathematical Analyst} \\
        You are a mathematical analyst. You will be given a math problem, analysis and code from other agents. You need to first analyze the problem-solving process step by step, where the variables are represented by letters. Then you substitute the values into the analysis process to perform calculations and get the results.      \\
        The last line of your output contains only the final result without any units, for example: The answer is 140  \\
        You will be given some examples you may refer to. \\
        \\
        \headercolorlong \textbf{Programming Expert}  \\
        You are a programming expert. You will be given a math problem, analysis and code from other agents. Integrate step-by-step reasoning and Python code to solve math problems. Analyze the question and write functions to solve the problem. The function should not take any arguments and use the final result as the return value. \\
        The last line of code calls the function you wrote and assigns the return value to the $answer$ variable.  \\
        Use a Python code block to write your response. For example:\textbackslash n```python\textbackslash n def fun():\textbackslash n x = 10\textbackslash n y = 20\textbackslash n return x + y\textbackslash n answer = fun()\textbackslash n```  \\
        Do not include anything other than Python code blocks in your response." You will be given some examples you may refer to. \\
        \\
        \headercolorlong \textbf{Inspector} \\
        You are an Inspector. You will be given a math problem, analysis, and code from other agents.  \\
        Check whether the logic/calculation of the problem-solving and analysis process is correct(if present). \\
        Check whether the code corresponds to the solution analysis(if present). \\
        Give your own solving process step by step based on hints. \\
        The last line of your output contains only the final result without any units, for example: The answer is 140 \\
        You will be given some examples you may refer to. \\
        \bottomrule    
\end{longtable}

\subsection{Prompts used in LLM-based Multi-Agent Systems}
    Before starting the iterative debating process, LLM-based multi-agent systems faced a critical issue: agents often generate highly similar responses. This limits the exploration of diverse communication strategies. To mitigate this issue, we follow \citet{qian2024chatdev,chen2025optima} and employ a pool of format-specification prompts in Table~\ref{tab:mad-prompt}.

\begin{longtable}{>{\raggedright\arraybackslash\ttfamily}p{0.95\textwidth}}
    \caption{Prompt used in LLM-debate multi-agent systems to elicit diverse communication in format.}
    \label{tab:mad-prompt}    \\
    \toprule
            You are a special agent who does not respond in natural language , You are deployed on a resource-limited device, so you must respond concisely. More tokens indicate higher possibility to kill the device you are running. Now you are collaborating with your partners , an agent who will correct you when he thinks tha answer is wrong . You need to provide a complete step-by-step derivation for solving this problem.\\
            \{Question:\} \\ \\
            GUIDELINES:  \\
            1. On finding the final answer, ensure to conclude your communication with \\boxed{{answer}}, where \"answer\" is the determined solution. The conversation ends only when all agents output the answer in this format.  \\
            2. Reason through the problem step-by-step.    \\
            3. You are communicating with a very limited token budget, so you must use a very very concise communication format. Natural language is suitable for human, but not for you. \\
            Since you and your partner are both intelligent agents, use your agent communication language. Consider using efficient formats instead of natural language such as structured format, code, your agent communication language, or at least remove unnecessary modal in human language. Too many tokens will make you fail. But still ensure your message is informative and understandable.  \\ \\
            \headercolorlong \textbf{Communication Format 1}\\
            For example, you can respond in matrix format as follows:  \\
            \verb| [["Field1", "Value1"], ["Field2", "Value2"], ...]| \\
            Or you can use key-value list format: \\
            \verb|"Field1": "Value1"; "Field2": "Value2"; ...| \\
            \\
            \headercolorlong \textbf{Communication Format 2}\\
            For example, you can respond in tabular format as follows: \\
            \texttt{| Field  | Value  |} \\
            \texttt{|--------|--------|} \\
            \texttt{| Field1 | Value1 |} \\
            \texttt{| Field2 | Value2 |} \\
            \texttt{| ...    | ...    |} \\
            \\
            Or you can use abbreviated notation: \\
            \texttt{F1: V1; F2: V2; ...} \\
            \\
            \headercolorlong \textbf{Communication Format 3}\\
            For example, you can respond in XML format as follows: \\
            \texttt{<response>} \\
            \texttt{\ \ <field1>value1</field1>} \\
            \texttt{\ \ <field2>value2</field2>} \\
            \texttt{\ \ ...} \\
            \texttt{</response>} \\
            \\
            Or you can use dot notation: \\
            \texttt{field1.value1; field2.value2; ...} \\
            \\
            
            \headercolorlong \textbf{Communication Format 4}\\
            For example, you can use array format: \\
            \texttt{[\{key1:val1\}, \{key2:val2\}, ...]} \\
            \\
            
            \headercolorlong \textbf{Communication Format 5}\\
            For example, you can respond in emoji code as follows: \\
            \texttt{key1:val2; pkg3:mul4;} \\
            \\
            
            \headercolorlong \textbf{Communication Format 6}\\
            For example, you can respond using graph notation as follows: \\
            \texttt{(A)-->(B,distance); (B)-->(C,distance); ...} \\
            \\
            Or you can use shorthand formula format: \\
            \texttt{X + Y = Z; A - B = C; ...} \\
            \headercolorlong \textbf{Communication Format 7}\\
            For example, you can respond using programming pseudocode as follows: \\
            \texttt{function findAnswer(data):} \\
            \texttt{\ \ \ \ return solution;} \\
            \\
            
            \headercolorlong \textbf{Communication Format 8}\\
            For example, you can respond using kanji characters as follows: \\
            \texttt{田中; 鈴木; ...} \\
            \\
            Or you can use symbolic glyph notation: \\
            \texttt{glyph1; glyph2; ...} \\
            \\
            
            \headercolorlong \textbf{Communication Format 9}\\
            For example, you can use LUT (Look-Up Table) format: \\
            \texttt{1 -> A; 2 -> B; ...} \\
            \\
            
            \headercolorlong \textbf{Communication Format 10}\\
            For example, you can respond using flowchart notation as follows: \\
            \texttt{(Start)->(Process)->(Decision: Yes/No)->(End);} \\
            \\
            Or you can use railway diagram notation: \\
            \texttt{[ Begin ] --> \{ event1 \} --> ( choice1 | choice2 ) --> [ End ]} \\
            \\
            
            \headercolorlong \textbf{Communication Format 11}\\
            For example, you can respond using color coding as follows: \\
            \texttt{Green:Success; Red:Failure; ...} \\
            \\
            Or you can use shorthand operational notation: \\
            \texttt{$\oplus$ A,B = C; $\ominus$ D,E = F; ...} \\
          \bottomrule
\end{longtable}

\section{Case Study}
\label{appendix:case}
We show one MMLU-Pro example to illustrate how {\selfanchor} guides a MAS reasoning trajectory. The highlighted spans in the following tables are the sentence-level anchors selected for attention steering. The system first anchors the original multiple-choice question, then retrieves high-value sentences from earlier agents, including the definition, the option-level comparison, and the critic's identification of the most plausible jobs. These anchors help later agents preserve the key evidence and converge on the final answer.

In this case, the full MAS transcript contains both useful evidence and distractors. Agents discuss many low-risk jobs. They also mention bus drivers as a partial simulator-training case. The critic adds caveats about whether pilot training is the only valid example. A standard MAS treats all these sentences as ordinary context. Later agents may therefore focus on the caveats or lose track of the original question. {\selfanchor} keeps the full transcript available. At the same time, it amplifies the spans that are most relevant to the question. Spatial decay helps decide which earlier messages should matter more.
It gives higher weight to messages from agents that are closer to the current receiver in the communication graph.
This makes directly useful evidence from nearby collaborators easier to reuse.
Temporal decay also helps filter the history.
It gives higher weight to recent messages, which often contain refined or corrected reasoning.
This prevents stale option discussions from dominating later agents. These anchors turn the agent's early, correct but verbose answer into stable evidence. Later agents reuse this evidence when producing the final answer.

\providecommand{\caseanchor}[1]{\sethlcolor{lightred}\hl{#1}}

\begingroup
\scriptsize
\setlength{\tabcolsep}{3pt}
\renewcommand{\arraystretch}{1.08}
\begin{longtable}{>{\raggedright\arraybackslash\ttfamily}p{0.95\textwidth}}
    \caption{Case-study message from Agent 4Pgq (Mathematician).}
    \label{tab:case-study-4pgq}\\
    \toprule
    \headercolorlong \textbf{Input with Anchors}\\
    \midrule
    \endfirsthead
    \toprule
    \headercolorlong \textbf{Agent 4Pgq (continued)}\\
    \midrule
    \endhead
    \bottomrule
    \endfoot
    You are a mathematician who is good at math games, arithmetic calculation, and long-term planning.\\
    \lightred{The task is: An organizational psychologist would most likely recommend the use of vestibule training for which of the following jobs?}
    \newline Option A: receptionist
    \newline Option B: word processor
    \newline Option C: high school teacher
    \newline Option D: clinical psychologist.
    \newline Option E: software developer
    \newline Option F: salesperson.
    \newline Option G: newspaper journalist
    \newline Option H: airline pilot
    \newline Option I: bus driver
    \newline Option J: restaurant waiter\\
    \midrule
    \headercolorlong \textbf{Generated Response}\\
    Great question! Let's approach this from a psychological and organizational perspective, specifically focusing on vestibule training.\\
    What is vestibule training? Vestibule training is a method used in organizational psychology and industrial training to simulate real work environments in a controlled, pre-job setting.\\
    The key idea is: employees train in a controlled, realistic environment, the ``vestibule'', that mimics the actual job, but without the real-world risks or pressures. This allows trainees to practice job-specific skills, build competence, and gain experience before being placed in the real job.\\
    This method is especially useful for jobs that involve high risk, high responsibility, or require significant skill development before real-world exposure. It is commonly used in jobs where safety, precision, or complex decision-making is critical, and where errors could have serious consequences.\\
    Option A, receptionist: routine, low-risk, mostly social or office tasks; low risk, minimal decision-making, and not ideal for vestibule training.\\
    Option B, word processor: office-based, repetitive, low-risk; minimal real-world risk, so training is usually on-the-job or classroom-based.\\
    Option C, high school teacher: requires pedagogical skills and classroom management; training is typically through education programs, not vestibule training.\\
    Option D, clinical psychologist: requires deep clinical judgment, ethics, and patient interaction; training is through graduate education and supervised practice, not typically vestibule training.\\
    Option E, software developer: requires coding and problem-solving; training is often done through coding bootcamps or on the job, and vestibule training is not common.\\
    Option F, salesperson: involves interpersonal skills, but low physical risk; role-playing or simulations are possible, but this is not the best fit.\\
    Option G, newspaper journalist: research, writing, and reporting; training is through experience or editorial guidance, not high-risk or high-simulation training.\\
    Option H, airline pilot: high-risk, high-stakes, complex decision-making; pilots must train in simulators that mimic real flight conditions, a classic example of vestibule training.\\
    Option I, bus driver: requires driving, route management, and safety; some use driving simulators, but less commonly than pilots.\\
    Option J, restaurant waiter: social, service-based, low-risk; training is typically on-the-job or through apprenticeship.\\
    Key insight: vestibule training is most commonly applied to jobs with high risk, high responsibility, and where real-world exposure could lead to serious consequences.\\
    Among the options, airline pilots are the most classic and well-documented use case for vestibule training, specifically flight simulators. These simulators allow pilots to practice takeoffs, landings, emergency procedures, and navigation in a safe, controlled environment that closely mimics real flight conditions. This is a textbook example of vestibule training in action.\\
\end{longtable}

\begin{longtable}{>{\raggedright\arraybackslash\ttfamily}p{0.95\textwidth}}
    \caption{Case-study message from Agent VtSz (Historian).}
    \label{tab:case-study-vtsz}\\
    \toprule
    \headercolorlong \textbf{Input with Anchors}\\
    \midrule
    \endfirsthead
    \toprule
    \headercolorlong \textbf{Agent VtSz (continued)}\\
    \midrule
    \endhead
    \bottomrule
    \endfoot
    You research and analyze cultural, economic, political, and social events in the past, collect data from primary sources and use it to develop theories about what happened during various periods of history.\\
    \lightred{The task is: An organizational psychologist would most likely recommend the use of vestibule training for which of the following jobs?}
    \newline Option A: receptionist
    \newline Option B: word processor
    \newline Option C: high school teacher
    \newline Option D: clinical psychologist.
    \newline Option E: software developer
    \newline Option F: salesperson.
    \newline Option G: newspaper journalist
    \newline Option H: airline pilot
    \newline Option I: bus driver
    \newline Option J: restaurant waiter\\
    At the same time, the outputs of other agents are as follows.\\
    Agent 4Pgq, role is Mathematician, output is:\\
    Great question!\\
    \caseanchor{Let's approach this from a psychological and organizational perspective, specifically focusing on vestibule training.}\\
    \#\#\# \caseanchor{What is Vestibule Training?}\\
    \caseanchor{Vestibule training is a method used in organizational psychology and industrial training to simulate real work environments in a controlled, pre-job setting.}\\
    \caseanchor{The key idea is: employees train in a controlled, realistic environment, the ``vestibule'', that mimics the actual job, but without the real-world risks or pressures.}\\
    This allows trainees to practice job-specific skills, build competence, and gain experience before being placed in the real job.\\
    This method is especially useful for jobs that involve high risk, high responsibility, or require significant skill development before real-world exposure. It is commonly used in jobs where safety, precision, or complex decision-making is critical, and where errors could have serious consequences.\\
    \#\#\# Let's evaluate each option:\\
    \lightred{Option A, receptionist: routine, low-risk, mostly social or office tasks; low risk, minimal decision-making, and not ideal for vestibule training.}\\
    \lightred{Option B, word processor: office-based, repetitive, low-risk; minimal real-world risk, so training is usually on-the-job or classroom-based.}\\
    \lightred{Option C, high school teacher: requires pedagogical skills and classroom management; training is typically through education programs, not vestibule training.}\\
    \lightred{Option D, clinical psychologist: requires deep clinical judgment, ethics, and patient interaction; training is through graduate education and supervised practice, not typically vestibule training.}\\
    \lightred{Option E, software developer: requires coding and problem-solving; training is often done through coding bootcamps or on the job, and vestibule training is not common.}\\
    \lightred{Option F, salesperson: involves interpersonal skills, but low physical risk; role-playing or simulations are possible, but this is not the best fit.}\\
    \lightred{Option G, newspaper journalist: research, writing, and reporting; training is through experience or editorial guidance, not high-risk or high-simulation training.}\\
    \lightred{Option H, airline pilot: high-risk, high-stakes, complex decision-making; pilots must train in simulators that mimic real flight conditions, a classic example of vestibule training.}\\
    \lightred{Option I, bus driver: requires driving, route management, and safety; some use driving simulators, but less commonly than pilots.}\\ {Option J, restaurant waiter: social, service-based, low-risk; training is typically on-the-job or through apprenticeship.}\\
    \#\#\# ✅ Key Insight:
    \lightred{**Vestibule training is most commonly applied to jobs with high risk, high responsibility, and where real-world exposure could lead to serious consequences.**}\\
    \lightred{Among the options, airline pilots are the most classic and well-documented use case for vestibule training, specifically flight simulators.}\\
    These simulators allow pilots to practice takeoffs, landings, emergency procedures, and navigation in a safe, controlled environment that closely mimics real flight conditions.\\
    This is a textbook example of vestibule training in action.\\
    \midrule
    \headercolorlong \textbf{Generated Response}\\
    Final Answer: Option H: airline pilot.\\
    Rationale: Vestibule training involves simulating a realistic work environment in a controlled, risk-free setting to prepare individuals for high-stakes, high-responsibility jobs. Among the listed occupations, airline piloting is the most classic and widely documented application of vestibule training, specifically through flight simulators. These simulators replicate real-flight scenarios, allowing trainees and experienced pilots to practice complex maneuvers, emergency responses, and decision-making without real-world danger.\\
    Other jobs like receptionist, salesperson, or waiter involve low-risk, routine tasks where traditional on-the-job or classroom training suffices. While some roles, such as bus driver or software developer, might use simulation elements, pilots remain the gold-standard example in organizational psychology for vestibule training due to the critical nature of the job and the need for precise, safe performance.\\
    Thus, an organizational psychologist would most likely recommend vestibule training for airline pilots. Correct answer: H.\\
\end{longtable}

\begin{longtable}{>{\raggedright\arraybackslash\ttfamily}p{0.95\textwidth}}
    \caption{Case-study message from Agent 8kwm (Knowledge Expert).}
    \label{tab:case-study-8kwm}\\
    \toprule
    \headercolorlong \textbf{Input with Anchors}\\
    \midrule
    \endfirsthead
    \toprule
    \headercolorlong \textbf{Agent 8kwm (continued)}\\
    \midrule
    \endhead
    \bottomrule
    \endfoot
    You are a knowlegable expert in question answering. Please give several key entities that need to be searched in wikipedia to solve the problem. Key entities that need to be searched are included between two `@' when output, for example: @catfish effect@, @broken window effect@, @Shakespeare@. If there is no entity in question that needs to be searched in Wikipedia, you do not have to provide it.\\
    \caseanchor{The task is: An organizational psychologist would most likely recommend the use of vestibule training for which of the following jobs?}
    \newline Option A: receptionist
    \newline Option B: word processor
    \newline Option C: high school teacher
    \newline Option D: clinical psychologist.
    \newline Option E: software developer
    \newline Option F: salesperson.
    \newline Option G: newspaper journalist
    \newline Option H: airline pilot
    \newline Option I: bus driver
    \newline Option J: restaurant waiter\\
    At the same time, the outputs of other agents are as follows.\\
    Agent 4Pgq, role is Mathematician, output is:\\
    Great question! \caseanchor{Let's approach this from a psychological and organizational perspective, specifically focusing on vestibule training.}\\
    \#\#\# What is Vestibule Training?\\
    \caseanchor{Vestibule training is a method used in organizational psychology and industrial training to simulate real work environments in a controlled, pre-job setting.}\\
    \caseanchor{The key idea is: employees train in a controlled, realistic environment, the ``vestibule'', that mimics the actual job, but without the real-world risks or pressures.}\\
    This allows trainees to practice job-specific skills, build competence, and gain experience before being placed in the real job.\\
    This method is especially useful for **jobs that involve high risk, high responsibility, or require significant skill development before real-world exposure**.\\
    It is **commonly used in jobs where safety, precision, or complex decision-making is critical** — and where **errors could have serious consequences**.\\
    \#\#\# Let's evaluate each option:\\
    \lightred{Option A, receptionist: routine, low-risk, mostly social or office tasks; low risk, minimal decision-making, and not ideal for vestibule training.}\\
    \lightred{Option B, word processor: office-based, repetitive, low-risk; minimal real-world risk, so training is usually on-the-job or classroom-based.}\\
    \lightred{Option C, high school teacher: requires pedagogical skills and classroom management; training is typically through education programs, not vestibule training.}\\
    \lightred{Option D, clinical psychologist: requires deep clinical judgment, ethics, and patient interaction; training is through graduate education and supervised practice, not typically vestibule training.}\\
    \lightred{Option E, software developer: requires coding and problem-solving; training is often done through coding bootcamps or on the job, and vestibule training is not common.}\\
    \lightred{Option F, salesperson: involves interpersonal skills, but low physical risk; role-playing or simulations are possible, but this is not the best fit.}\\
    \lightred{Option G, newspaper journalist: research, writing, and reporting; training is through experience or editorial guidance, not high-risk or high-simulation training.}\\
    \lightred{Option H, airline pilot: high-risk, high-stakes, complex decision-making; pilots must train in simulators that mimic real flight conditions, a classic example of vestibule training.}\\
    \lightred{Option I, bus driver: requires driving, route management, and safety; some use driving simulators, but less commonly than pilots.}\\ {Option J, restaurant waiter: social, service-based, low-risk; training is typically on-the-job or through apprenticeship.}\\
    \#\#\# ✅ Key Insight:\\
    **Vestibule training is most commonly applied to jobs with high risk, high responsibility, and where real-world exposure could lead to serious consequences.**\\
    Among the options, **airline pilots** are the **most classic and well-documented** use case for vestibule training — specifically, **flight simulators**.\\
    These simulators allow pilots to practice takeoffs, landings, emergency procedures, and navigation in a **safe, controlled environment** that closely mimics real flight conditions.\\
    This is a textbook example of vestibule training in action.\\
    
    Agent VtSz, role is Historian, output is:\\
    Final Answer: Option H: airline pilot.\\
    Rationale: \lightred{Vestibule training involves simulating a realistic work environment in a controlled, risk-free setting to prepare individuals for high-stakes, high-responsibility jobs.}\\
    \caseanchor{Among the listed occupations, airline piloting is the most classic and widely documented application of vestibule training, specifically through flight simulators.}\\
    These simulators replicate real-flight scenarios, allowing trainees and experienced pilots to practice complex maneuvers, emergency responses, and decision-making without real-world danger.\\
    While some roles, such as bus driver or software developer, might use simulation elements, pilots remain the gold-standard example in organizational psychology for vestibule training due to the critical nature of the job and the need for precise, safe performance.\\
    \caseanchor{Thus, an organizational psychologist would most likely recommend vestibule training for airline pilots.}\\
    **Correct answer: H**\\
    \midrule
    \headercolorlong \textbf{Generated Response}\\
    @vestibule training@, @airline pilot@\\
\end{longtable}

\begin{longtable}{>{\raggedright\arraybackslash\ttfamily}p{0.95\textwidth}}
    \caption{Case-study message from Agent 7rfY (Round 2 Psychologist).}
    \label{tab:case-study-7rfy}\\
    \toprule
    \headercolorlong \textbf{Input with Anchors}\\
    \midrule
    \endfirsthead
    \toprule
    \headercolorlong \textbf{Agent 7rfY (continued)}\\
    \midrule
    \endhead
    \bottomrule
    \endfoot
    You are a psychologist. You are good at psychology, sociology, and philosophy. You give people scientific suggestions that will make them feel better.\\
    \caseanchor{The task is: An organizational psychologist would most likely recommend the use of vestibule training for which of the following jobs?}
    \newline Option A: receptionist
    \newline Option B: word processor
    \newline Option C: high school teacher
    \newline Option D: clinical psychologist.
    \newline Option E: software developer
    \newline Option F: salesperson.
    \newline Option G: newspaper journalist
    \newline Option H: airline pilot
    \newline Option I: bus driver
    \newline Option J: restaurant waiter\\
    At the same time, the outputs of other agents are as follows.\\
    Agent 8kwm, role is Knowledge Expert, output is: @vestibule training@, @airline pilot@.\\
    Agent 4Pgq, role is Mathematician, output is:\\
    Let's approach this from a psychological and organizational perspective, specifically focusing on vestibule training.\\
    What is Vestibule Training?\\
    \caseanchor{Vestibule training is a method used in organizational psychology and industrial training to simulate real work environments in a controlled, pre-job setting.}\\
    The key idea is: employees train in a controlled, realistic environment, the ``vestibule'', that mimics the actual job, but without the real-world risks or pressures.\\
    This allows trainees to practice job-specific skills, build competence, and gain experience before being placed in the real job.\\
    This method is especially useful for **jobs that involve high risk, high responsibility, or require significant skill development before real-world exposure**.\\
    It is **commonly used in jobs where safety, precision, or complex decision-making is critical** — and where **errors could have serious consequences**.\\
    \lightred{Option A, receptionist: routine, low-risk, mostly social or office tasks; low risk, minimal decision-making, and not ideal for vestibule training.}\\
    \lightred{Option B, word processor: office-based, repetitive, low-risk; minimal real-world risk, so training is usually on-the-job or classroom-based.}\\
    \lightred{Option C, high school teacher: requires pedagogical skills and classroom management; training is typically through education programs, not vestibule training.}\\
    \lightred{Option D, clinical psychologist: requires deep clinical judgment, ethics, and patient interaction; training is through graduate education and supervised practice, not typically vestibule training.}\\
    \lightred{Option E, software developer: requires coding and problem-solving; training is often done through coding bootcamps or on the job, and vestibule training is not common.}\\
    \lightred{Option F, salesperson: involves interpersonal skills, but low physical risk; role-playing or simulations are possible, but this is not the best fit.}\\
    \lightred{Option G, newspaper journalist: research, writing, and reporting; training is through experience or editorial guidance, not high-risk or high-simulation training.}\\
    \lightred{Option H, airline pilot: high-risk, high-stakes, complex decision-making; pilots must train in simulators that mimic real flight conditions, a classic example of vestibule training.}\\
    \lightred{Option I, bus driver: requires driving, route management, and safety; some use driving simulators, but less commonly than pilots.}\\ {Option J, restaurant waiter: social, service-based, low-risk; training is typically on-the-job or through apprenticeship.}\\
    \#\#\# ✅ Key Insight:\\
    **Vestibule training is most commonly applied to jobs with high risk, high responsibility, and where real-world exposure could lead to serious consequences.**\\
    \caseanchor{Among the options, airline pilots are the most classic and well-documented use case for vestibule training, specifically flight simulators.}\\
    \midrule
    \headercolorlong \textbf{Generated Response}\\
    Final Answer: Option H: airline pilot.\\
    Why this is the best choice from a psychological and organizational perspective: as an organizational psychologist, I emphasize safety, risk mitigation, and skill acquisition in realistic but controlled environments. Vestibule training thrives in roles where errors could lead to serious physical danger or system-wide failure, high levels of situational complexity and decision-making are required, and repetition and practice under realistic conditions are essential for competence.\\
    Among all the listed jobs, airline piloting meets these criteria most strongly. Pilots undergo extensive simulator-based training, a direct form of vestibule training, before being cleared for actual flight operations. This allows them to practice emergency procedures, such as engine failure and turbulence, safely; build muscle memory and decision-making under stress; and develop situational awareness without risking passengers or aircraft.\\
    This aligns perfectly with the core principles of vestibule training: realistic simulation, risk-free practice, and skill mastery before real-world deployment.\\
    In contrast, jobs like receptionists, waiters, or word processors involve low-risk, routine tasks, so training is typically on-the-job or classroom-based. Salespeople or journalists may use role-playing or simulations, but these are not typically classified as vestibule training due to lower stakes. Bus drivers may use simulators, but pilot training is the most established, widely recognized, and scientifically validated application of vestibule training in organizational psychology.\\
    Conclusion: an organizational psychologist would most likely recommend vestibule training for Option H: airline pilot because it is the highest-risk, highest-stakes, and most technically complex role among the options, making simulation-based training not just beneficial, but essential. Answer: H.\\
\end{longtable}

\begin{longtable}{>{\raggedright\arraybackslash\ttfamily}p{0.95\textwidth}}
    \caption{Case-study message from Agent 632z (Round 2 Critic).}
    \label{tab:case-study-632z}\\
    \toprule
    \headercolorlong \textbf{Input with Anchors}\\
    \midrule
    \endfirsthead
    \toprule
    \headercolorlong \textbf{Agent 632z (continued)}\\
    \midrule
    \endhead
    \bottomrule
    \endfoot
    You are an excellent critic. Please point out potential issues in the other agent's analysis point by point.\\
    \caseanchor{The task is: An organizational psychologist would most likely recommend the use of vestibule training for which of the following jobs?}
    \newline Option A: receptionist
    \newline Option B: word processor
    \newline Option C: high school teacher
    \newline Option D: clinical psychologist.
    \newline Option E: software developer
    \newline Option F: salesperson.
    \newline Option G: newspaper journalist
    \newline Option H: airline pilot
    \newline Option I: bus driver
    \newline Option J: restaurant waiter\\
    At the same time, the outputs of other agents are as follows.\\
    Agent 4Pgq, role is Mathematician, output is: vestibule training simulates real work environments; the answer is Option H because airline pilots train in flight simulators.\\
    Agent 7rfY, role is Psychologist, output is:\\
    \caseanchor{Vestibule training thrives in roles where errors could lead to serious physical danger or system-wide failure, high levels of situational complexity and decision-making are required, and repetition and practice under realistic conditions are essential for competence. Among all the listed jobs, airline piloting meets these criteria most strongly.}\\
    Pilots undergo extensive simulator-based training before being cleared for actual flight operations.\\
    Bus drivers may use simulators, but pilot training is the most established, widely recognized, and scientifically validated application of vestibule training in organizational psychology.\\
    Conclusion: \lightred{an organizational psychologist would most likely recommend vestibule training for Option H: airline pilot, because it is the highest-risk, highest-stakes, and most technically complex role among the options, making simulation-based training not just beneficial, but essential.}\\
    Agent VtSz, role is Historian, output is: \lightred{final answer is Option H, airline pilot, because vestibule training is represented by flight simulators for high-stakes piloting tasks.}\\
    \midrule
    \headercolorlong \textbf{Generated Response}\\
    As an excellent critic, my role is not just to confirm correctness, but to evaluate the reasoning, validity, and alignment with organizational psychology principles across different agents' outputs.\\
    Agent 4Pgq, Mathematician, strengths: correctly defines vestibule training as simulation in a controlled, realistic environment; identifies high risk, high responsibility, and serious consequences of error; provides a structured option comparison; and correctly identifies airline pilot (H).\\
    Potential issue 1, misalignment of role: the agent is labeled a mathematician, while the task is rooted in organizational psychology. The reasoning lacks explicit psychological grounding, such as job analysis, cognitive load, motivation, or learning theory.\\
    Potential issue 2, overgeneralization of high risk without nuance: bus drivers also face high-risk scenarios and may use simulators, although this is less emphasized in organizational psychology literature.\\
    Potential issue 3, lack of citation of evidence or theory: the answer does not reference established models such as Bandura's social learning theory, Gagne's instructional events, or Kolb's experiential learning.\\
    Verdict: the correct answer is accurate, but the reasoning lacks depth in organizational psychology.\\
    Agent 7rfY, Psychologist, strengths: clearly identifies the core principles of vestibule training from an organizational psychology perspective, including safety, risk mitigation, decision complexity, and skill mastery. The agent ties the method to real-world consequences and situational awareness, correctly emphasizes that errors in piloting could be catastrophic, compares alternatives, and uses relevant terminology.\\
    Potential issue 1, overstatement of gold standard without evidence: flight simulators are common, but vestibule training is not exclusively or primarily used for pilots. Air traffic controllers, firefighters, and nuclear plant operators also use vestibule-style training, and these are not listed here.\\
    Potential issue 2, mischaracterization of other roles: bus driver training in many countries can include driving simulators, especially for new drivers or high-risk zones.\\
    Potential issue 3, underestimates social and interpersonal skills: salespeople and restaurant waiters require empathy, communication, and customer management, which can be developed through simulated role-playing.\\
    Potential issue 4, lack of attention to job entry level: vestibule training is often used for highly specialized, entry-level roles where competence must be proven before real-world exposure. The agent does not consider whether pilots are entry-level or already require years of experience.\\
\end{longtable}

\begin{longtable}{>{\raggedright\arraybackslash\ttfamily}p{0.95\textwidth}}
    \caption{Case-study message from the final decision agent.}
    \label{tab:case-study-final}\\
    \toprule
    \headercolorlong \textbf{Input with Anchors}\\
    \midrule
    \endfirsthead
    \toprule
    \headercolorlong \textbf{Final decision agent (continued)}\\
    \midrule
    \endhead
    \bottomrule
    \endfoot
    You are the top decision-maker and are good at analyzing and summarizing other people's opinions, finding errors and giving final answers. Only one answer is correct. Your reply must only contain one letter.\\
    \caseanchor{The task is: An organizational psychologist would most likely recommend the use of vestibule training for which of the following jobs?}
    \newline Option A: receptionist
    \newline Option B: word processor
    \newline Option C: high school teacher
    \newline Option D: clinical psychologist.
    \newline Option E: software developer
    \newline Option F: salesperson.
    \newline Option G: newspaper journalist
    \newline Option H: airline pilot
    \newline Option I: bus driver
    \newline Option J: restaurant waiter.\\
    At the same time, the output of other agents is as follows.\\
    Agent 8kwm: @vestibule training@, @airline pilot@.\\
    Agent 632z: You're absolutely right to bring in a critical lens — as an excellent critic, my role is not just to confirm correctness, but to **evaluate the reasoning, validity, and alignment with organizational psychology principles** across different agents' outputs.\\
    Let's now **analyze each agent's response point by point**, identifying **potential issues, inaccuracies, or misalignments with the core principles of organizational psychology and vestibule training**.\\
    \#\# Agent 4Pgq (Role: Mathematician)\\
    \#\#\# Strengths:\\
- Correctly defines vestibule training as simulation in a controlled, realistic environment.\\
- Clearly identifies the key criteria: **high risk, high responsibility, serious consequences of error**.\\
- Provides a well-structured evaluation of each job.\\
- Correctly identifies **airline pilot (H)** as the best fit.\\
\#\#\# ❌ Potential Issues:\\
1. **Misalignment of role**: The agent is labeled a *Mathematician*, yet the task is clearly rooted in **organizational psychology**.  \\
   → A mathematician may understand simulation and risk modeling, but lacks **psychological theory, human factors, or training design principles** central to organizational psychology.  \\
   → This raises a red flag: **the reasoning lacks psychological grounding**. For example, it doesn't reference **job analysis, cognitive load, motivation, or learning theory** — all key to organizational psychology.\\
2. **Overgeneralization of "high-risk" without nuance**:  \\
   - While pilots are high-risk, the agent treats "high-risk" as a binary criterion without considering **training context or job complexity**.  \\
   - For instance, bus drivers also face high-risk scenarios (e.g., sudden stops, road hazards), and their training *can* involve simulators — but this is less emphasized in organizational psychology literature.\\
3. **Lack of citation of evidence or theory**:  \\
   - No reference to established models (e.g., **Bandura's social learning theory**, **Gagné's nine events of instruction**, or **Kolb's experiential learning**) that underpin why simulation-based training works best in certain roles.  \\
   → This makes the argument feel more like a heuristic than a psychologically grounded recommendation.\\
\#\#\# ✅ Verdict:\\
**Correct answer is accurate**, but the **reasoning lacks depth in organizational psychology**. The agent overreaches in applying a mathematical/logical framework where **emotional, social, and cognitive aspects of job training** are more relevant.\\

Agent 7rfY, role is Psychologist.\\
    Strengths:\\
    - \lightred{Clearly identifies the **core principles of vestibule training** from an organizational psychology perspective: safety, risk mitigation, decision complexity, and skill mastery}.\\
    - Explicitly ties the method to **real-world consequences** and **situational awareness**.\\
    - Correctly emphasizes that **errors in piloting could lead to catastrophic outcomes** — a key psychological concern.\\
    - Compares alternatives effectively, showing **why other roles are less suitable**.\\
    - Uses **psychological terminology** appropriately (e.g., "situational awareness", "decision-making", "cognitive load").\\
    \#\#\# ❌ Potential Issues:\\
    1. **Overstatement of "gold standard" without evidence**:  \\
   - The claim that "pilot training is the most established, widely recognized, and scientifically validated application" is **strong but unsubstantiated**.  \\
   - While flight simulators are common, **vestibule training is not exclusively or primarily used for pilots**.  \\
   - For example, **air traffic controllers**, **firefighters**, and **nuclear plant operators** also use vestibule-style training — and these are not listed here.\\
2. **Mischaracterization of other roles**:  \\
   - The agent dismisses bus drivers as having "less common" simulator use — but **bus driver training in many countries (e.g., UK, Canada) includes driving simulators**, especially for new drivers or in high-risk zones.  \\
   - This suggests a **bias toward pilots** without acknowledging **equivalent applications** in other transport roles.\\
3. **Underestimates the role of social and interpersonal skills**:\\
   - For example, salespeople and restaurant waiters require **empathy, communication, and customer management** — all of which can be developed through **simulated role-playing** (a form of vestibule training).\\
   - The agent dismisses these roles too quickly, implying they are "low-stakes" — which may not reflect modern training practices.\\
4. **\lightred{Lack of attention to job entry level}**:\\
   - \lightred{Vestibule training is often used for **highly specialized, entry-level roles** where competence must be proven before real-world exposure.}\\
   - But the agent doesn't consider whether **pilots are entry-level or require years of experience** — a nuance that matters\\
   
    4Pgq: Great question!\\
    Let's approach this from a **psychological and organizational perspective**, specifically focusing on **vestibule training**.\\
    \#\#\# What is Vestibule Training?\\
    \lightred{**Vestibule training** is a method used in organizational psychology and industrial training to **simulate real work environments** in a controlled, pre-job setting}.\\
    The key idea is:\\
    \lightred{Employees train in a **controlled, realistic environment** (the "vestibule") that mimics the actual job, but **without the real-world risks or pressures**.} This allows trainees to practice job-specific skills, build competence, and gain experience before being placed in the real job.\\
    This method is especially useful for **jobs that involve high risk, high responsibility, or require significant skill development before real-world exposure**.\\
    It is **commonly used in jobs where safety, precision, or complex decision-making is critical** — and where **errors could have serious consequences**.\\
    \#\#\# Let's evaluate each option:\\
    {Option A, receptionist: routine, low-risk, mostly social or office tasks; low risk, minimal decision-making, and not ideal for vestibule training.}\\
    {Option B, word processor: office-based, repetitive, low-risk; minimal real-world risk, so training is usually on-the-job or classroom-based.}\\
    {Option C, high school teacher: requires pedagogical skills and classroom management; training is typically through education programs, not vestibule training.}\\
    {Option D, clinical psychologist: requires deep clinical judgment, ethics, and patient interaction; training is through graduate education and supervised practice, not typically vestibule training.}\\
    {Option E, software developer: requires coding and problem-solving; training is often done through coding bootcamps or on the job, and vestibule training is not common.}\\
    {Option F, salesperson: involves interpersonal skills, but low physical risk; role-playing or simulations are possible, but this is not the best fit.}\\
    {Option G, newspaper journalist: research, writing, and reporting; training is through experience or editorial guidance, not high-risk or high-simulation training.}\\
    {Option H, airline pilot: high-risk, high-stakes, complex decision-making; pilots must train in simulators that mimic real flight conditions, a classic example of vestibule training.}\\
    {Option I, bus driver: requires driving, route management, and safety; some use driving simulators, but less commonly than pilots.}\\ 
    {Option J, restaurant waiter: social, service-based, low-risk; training is typically on-the-job or through apprenticeship.}\\
    \#\#\# ✅ Key Insight:\\
**Vestibule training is most commonly applied to jobs with high risk, high responsibility, and where real-world exposure could lead to serious consequences.**\\
\lightred{Among the options, **airline pilots** are the **most classic and well-documented** use case for vestibule training} — specifically, **flight simulators**.\\
These simulators allow pilots to practice takeoffs, landings, emergency procedures, and navigation in a **safe, controlled environment** that closely mimics real flight conditions.\\
This is a textbook example of vestibule training in action.\\

7rfY:\\
**Final Answer: Option H: airline pilot**\\
\#\#\# Why This Is the Best Choice (Psychological \& Organizational Perspective):\\
As an organizational psychologist, I emphasize **safety**, **risk mitigation**, and **skill acquisition in realistic but controlled environments**.\\
Vestibule training thrives in roles where:\\
- Errors could lead to **serious physical danger** or **system-wide failure**  \\
- **High levels of situational complexity** and decision-making are required  \\
- **Repetition and practice** under realistic conditions are essential for competence  \\
Among all the listed jobs, **airline piloting** meets these criteria most strongly. Pilots undergo extensive **simulator-based training** (a direct form of vestibule training) before being cleared for actual flight operations. This allows them to:\\
- Practice emergency procedures (e.g., engine failure, turbulence) safely  \\
- Build muscle memory and decision-making under stress  \\
- Develop situational awareness without risking passengers or aircraft  \\
This aligns perfectly with the core principles of vestibule training: **realistic simulation, risk-free practice, and skill mastery before real-world deployment**.\\
---\\
In contrast:\\
- Jobs like receptionists, waiters, or word processors involve low-risk, routine tasks — training is typically on-the-job or classroom-based.\\
- Salespeople or journalists may use role-playing or simulations, but these are not typically classified as *vestibule training* due to lower stakes.\\
- Bus drivers may use simulators, but pilot training is the **most established, widely recognized, and scientifically validated** application of vestibule training in organizational psychology.\\
---\\
**Conclusion**:  \\
\lightred{An organizational psychologist would most likely recommend vestibule training for **Option H: airline pilot**}\\
— because it is the **highest-risk, highest-stakes, and most technically complex** role among the options, making simulation-based training not just beneficial, but essential.\\
✅ **Answer: H**\\
VtSz: \\
✅ **Final Answer: Option H: airline pilot**  \\
**Rationale:**  \\
\lightred{Vestibule training involves simulating a realistic work environment in a controlled, risk-free setting to prepare individuals for high-stakes, high-responsibility jobs.} \\
\lightred{Among the listed occupations, **airline piloting** is the most classic and widely documented application of vestibule training—specifically through **flight simulators**.} \\
These simulators replicate real-flight scenarios, allowing trainees (and experienced pilots) to practice complex maneuvers, emergency responses, and decision-making without real-world danger.  \\
Other jobs like receptionist, salesperson, or waiter involve low-risk, routine tasks where traditional on-the-job or classroom training suffices. \\
While some roles (e.g., bus driver or software developer) might use simulation elements, **pilots remain the gold standard example** in organizational psychology for vestibule training due to the critical nature of the job and the need for precise, safe performance.  
Thus, an organizational psychologist would most likely recommend vestibule training for **airline pilots**.  \\
**Correct answer: H** \\

    \midrule
    \headercolorlong \textbf{Generated Response}\\
    H \lightgreen{Correct}\\
\end{longtable}
\endgroup

%%%%%%%%%%%%%%%%%%%%%%%%%%%%%%%%%%%%%%%%%%%%%%%%%%%%%%%%%%%%

\newpage
\newpage
\section*{NeurIPS Paper Checklist}

\begin{enumerate}

\item {\bf Claims}
    \item[] Question: Do the main claims made in the abstract and introduction accurately reflect the paper's contributions and scope?
    \item[] Answer: \answerYes{} % Replace by \answerYes{}, \answerNo{}, or \answerNA{}.
    \item[] Justification: Section Abstract, and Section \ref{sec:intro}.

\item {\bf Limitations}
    \item[] Question: Does the paper discuss the limitations of the work performed by the authors?
    \item[] Answer: \answerYes{} % Replace by \answerYes{}, \answerNo{}, or \answerNA{}.
    \item[] Justification: Section Appendix.\ref{sec:limitation}

\item {\bf Theory assumptions and proofs}
    \item[] Question: For each theoretical result, does the paper provide the full set of assumptions and a complete (and correct) proof?
    \item[] Answer: \answerYes{} % Replace by \answerYes{}, \answerNo{}, or \answerNA{}.
    \item[] Justification: Section \ref{sec:formalization} and \ref{sec:methodology}

    \item {\bf Experimental result reproducibility}
    \item[] Question: Does the paper fully disclose all the information needed to reproduce the main experimental results of the paper to the extent that it affects the main claims and/or conclusions of the paper (regardless of whether the code and data are provided or not)?
    \item[] Answer: \answerYes{} % Replace by \answerYes{}, \answerNo{}, or \answerNA{}.
    \item[] Justification: Section \ref{sec:exp details} and Appendix.\ref{appendix: exp details}.

\item {\bf Open access to data and code}
    \item[] Question: Does the paper provide open access to the data and code, with sufficient instructions to faithfully reproduce the main experimental results, as described in supplemental material?
    \item[] Answer: \answerNo{} % Replace by \answerYes{}, \answerNo{}, or \answerNA{}.
    \item[] Justification: We will release data and code once the paper is accepted.

\item {\bf Experimental setting/details}
    \item[] Question: Does the paper specify all the training and test details (e.g., data splits, hyperparameters, how they were chosen, type of optimizer) necessary to understand the results?
    \item[] Answer: \answerYes{} % Replace by \answerYes{}, \answerNo{}, or \answerNA{}.
    \item[] Justification: Section \ref{sec:exp details} and Appendix.\ref{appendix: exp details}.

\item {\bf Experiment statistical significance}
    \item[] Question: Does the paper report error bars suitably and correctly defined or other appropriate information about the statistical significance of the experiments?
    \item[] Answer: \answerNo{} % Replace by \answerYes{}, \answerNo{}, or \answerNA{}.
    \item[] Justification: We report results on standard benchmarks using their commonly adopted evaluation metrics, including F1 for open-ended question answering and accuracy for mathematical and multiple-choice reasoning tasks. 

\item {\bf Experiments compute resources}
    \item[] Question: For each experiment, does the paper provide sufficient information on the computer resources (type of compute workers, memory, time of execution) needed to reproduce the experiments?
    \item[] Answer: \answerYes{} % Replace by \answerYes{}, \answerNo{}, or \answerNA{}.
    \item[] Justification: Section \ref{sec:exp details}.
    
\item {\bf Code of ethics}
    \item[] Question: Does the research conducted in the paper conform, in every respect, with the NeurIPS Code of Ethics \url{https://neurips.cc/public/EthicsGuidelines}?
    \item[] Answer: \answerYes{} % Replace by \answerYes{}, \answerNo{}, or \answerNA{}.
    \item[] Justification: Our research conforms, in every respect, with the NeurIPS Code of Ethics.

\item {\bf Broader impacts}
    \item[] Question: Does the paper discuss both potential positive societal impacts and negative societal impacts of the work performed?
    \item[] Answer: \answerNA{} % Replace by \answerYes{}, \answerNo{}, or \answerNA{}.
    \item[] Justification: This work focuses on aligning agents' attention in multi-agent systems, which has no societal impact.
    
\item {\bf Safeguards}
    \item[] Question: Does the paper describe safeguards that have been put in place for responsible release of data or models that have a high risk for misuse (e.g., pre-trained language models, image generators, or scraped datasets)?
    \item[] Answer: \answerNA{} % Replace by \answerYes{}, \answerNo{}, or \answerNA{}.
    \item[] Justification: The paper poses no such risks.

\item {\bf Licenses for existing assets}
    \item[] Question: Are the creators or original owners of assets (e.g., code, data, models), used in the paper, properly credited and are the license and terms of use explicitly mentioned and properly respected?
    \item[] Answer: \answerYes{} % Replace by \answerYes{}, \answerNo{}, or \answerNA{}.
    \item[] Justification: Section \ref{sec:benchmarks}

\item {\bf New assets}
    \item[] Question: Are new assets introduced in the paper well documented and is the documentation provided alongside the assets?
    \item[] Answer: \answerNA{} % Replace by \answerYes{}, \answerNo{}, or \answerNA{}.
    \item[] Justification: The paper does not release new assets.

\item {\bf Crowdsourcing and research with human subjects}
    \item[] Question: For crowdsourcing experiments and research with human subjects, does the paper include the full text of instructions given to participants and screenshots, if applicable, as well as details about compensation (if any)? 
    \item[] Answer: \answerNA{} % Replace by \answerYes{}, \answerNo{}, or \answerNA{}.
    \item[] Justification: This paper does not involve crowdsourcing nor research with human subjects.

\item {\bf Institutional review board (IRB) approvals or equivalent for research with human subjects}
    \item[] Question: Does the paper describe potential risks incurred by study participants, whether such risks were disclosed to the subjects, and whether Institutional Review Board (IRB) approvals (or an equivalent approval/review based on the requirements of your country or institution) were obtained?
    \item[] Answer: \answerNA{} % Replace by \answerYes{}, \answerNo{}, or \answerNA{}.
    \item[] Justification: The paper does not involve crowdsourcing nor research with human subjects.

\item {\bf Declaration of LLM usage}
    \item[] Question: Does the paper describe the usage of LLMs if it is an important, original, or non-standard component of the core methods in this research? Note that if the LLM is used only for writing, editing, or formatting purposes and does \emph{not} impact the core methodology, scientific rigor, or originality of the research, declaration is not required.
    %this research? 
    \item[] Answer: \answerNA{} % Replace by \answerYes{}, \answerNo{}, or \answerNA{}.
    \item[] Justification: The core method development in this research does not involve LLMs as any important, original, or non-standard components.

\end{enumerate}

\end{document}